\documentclass[preprint,12pt]{elsarticle}

\usepackage{amssymb}
\usepackage{amsmath}
\usepackage{float}
\usepackage{booktabs}
\usepackage{multirow}
\usepackage{subcaption}
\usepackage{threeparttable}
\usepackage{enumitem}
\usepackage[figuresright]{rotating}
\usepackage[colorlinks=true, citecolor=blue]{hyperref}

\def\nno{\nonumber}

\usepackage{xcolor}
\renewcommand{\(}{\left(}\renewcommand{\)}{\right)}

\newtheorem{remark}{Remark}

\newtheorem{proposition}{Proposition}

\journal{Neurocomputing}

\begin{document}

\begin{frontmatter}



\title{SigMA: Path Signatures and Multi-head Attention for Learning Parameters in fBm-driven SDEs} 


\author[1]{Xianglin Wu\corref{cor1}}
\ead{1210202Z1003@smail.swufe.edu.cn}
\cortext[cor1]{Corresponding author}
\author[2]{Chiheb Ben Hammouda}
\author[2]{Cornelis W. Oosterlee}

\affiliation[1]{
	organization={School of Mathematics, Southwestern University of Finance and Economics},
	city={Chengdu},
	postcode={611130},
	country={China}}
\affiliation[2]{
	organization={Mathematical Institute, Utrecht University},
	city={Utrecht},
	postcode={3584 CD},
	country={The Netherlands}}

\begin{abstract}
Stochastic differential equations (SDEs) driven by fractional Brownian motion (fBm) are increasingly used to model systems with rough dynamics and long-range dependence, such as those arising in quantitative finance and reliability engineering. However, these processes are non-Markovian and lack a semimartingale structure, rendering many classical parameter estimation techniques inapplicable or computationally intractable beyond very specific cases. This work investigates two central questions: (i) whether integrating path signatures into deep learning architectures can improve the trade-off between estimation accuracy and model complexity, and (ii) what constitutes an effective architecture for leveraging signatures as feature maps.

We introduce SigMA (Signature Multi-head Attention), a neural architecture that integrates path signatures with multi-head self-attention, supported by a convolutional preprocessing layer and a multilayer perceptron for effective feature encoding.  SigMA learns model parameters from synthetically generated paths of fBm-driven SDEs, including fractional Brownian motion, fractional Ornstein–Uhlenbeck, and rough Heston models, with a particular focus on estimating the Hurst parameter and on joint multi-parameter inference, and it generalizes robustly to unseen trajectories.   Extensive experiments on synthetic data and two real-world datasets (i.e., equity-index realized volatility and Li-ion battery degradation) show that SigMA consistently outperforms  CNN, LSTM, vanilla Transformer, and Deep Signature baselines in accuracy, robustness, and model compactness.  These results demonstrate that combining signature transforms with attention-based architectures provides an effective and scalable framework for parameter inference in stochastic systems with rough or persistent temporal structure.
\end{abstract}



\begin{keyword}
fractional Brownian motion \sep stochastic differential equations \sep non-Markovian processes \sep parameter estimation \sep path signatures \sep self-attention
\end{keyword}

\end{frontmatter}


\section{Introduction}
\label{sec-introduction}
Stochastic differential equations (SDEs) driven by fractional Brownian motion (fBm) arise in domains characterized by long-range dependence as well as pathwise roughness. Such applications include rough volatility modeling in quantitative finance \cite{gatheral2018volatility} and degradation dynamics in engineering systems such as Li-ion batteries \cite{shao2021degradation, boros2024deep}. A central quantity in models driven by fBm is the Hurst parameter $H \in (0,1)$, which    governs temporal memory, self-similarity, and the local regularity of sample paths. Accurate estimation of this parameter from observed time series is critical, but challenging due to the non-Markovian nature of fBm-driven models. Moreover, in many applications, multiple parameters must be estimated jointly, such as the Hurst exponent, volatility of volatility, correlation, and/or mean-reversion parameters in rough volatility models. This is especially relevant in settings where pricing and risk sensitivities depend on combinations of these parameters \cite{bayer2016pricing, el2019characteristic}.

For $H \neq \tfrac{1}{2}$, fBm-driven processes are inherently non-Markovian and do not admit a semimartingale structure. As a result, many classical parameter estimation techniques become inapplicable or computationally intractable beyond very specific settings. These challenges motivate the development of data-driven architectures that can extract path-level features robustly across a wide range of parameter regimes. Recent machine learning techniques, including deep feedforward neural networks (FNNs) \cite{mukherjee2023hurst}, convolutional neural networks (CNNs) \cite{stone2020calibrating}, and long short-term memory (LSTM) \cite{boros2024deep}, have been proposed to address these limitations. However, these approaches often lack robustness across different path regimes or do not exploit the intrinsic geometric structure of long-memory paths.

Path signatures, arising from rough path theory \cite{lyons2014rough}, offer a principled way to represent continuous-time trajectories in a coordinate-free manner.  Signature features have been shown to form an expressive and universal representation for path-dependent functionals, and have been used successfully in classification, forecasting, and medical time series \cite{Lyons_2019_DeepSignature, morrill2021neural}. Recent efforts have also combined signatures with transformer architectures for time series modeling \cite{moreno2024rough}. However, existing approaches typically focus on prediction or classification, rather than on parameter inference in fractional stochastic models.

In this work, we investigate two main questions in the context of fBm-driven processes:
\begin{itemize}
	\item whether integrating path signatures into deep learning architectures improves the trade-off between estimation accuracy and model complexity, and
	\item  what constitutes an effective architecture for leveraging signatures as feature maps.
\end{itemize}

To this end, we propose \textbf{SigMA} (Signature Multi-head Attention), a neural architecture that integrates path signature features with multi-head self-attention, enhanced by convolutional preprocessing and multilayer perceptron (MLP) layers. Unlike prior methods such as Deep Signature Networks \cite{Lyons_2019_DeepSignature}, LSTM-based estimators \cite{boros2024deep}, or vanilla Transformers \cite{vaswani_2017}, SigMA is specifically designed for parameter estimation in fBm-driven SDEs, across both rough and long-memory regimes.

We evaluate SigMA on both synthetic and empirical datasets. Synthetic experiments consider parameter estimation from simulated paths of fBm, fractional Ornstein–Uhlenbeck (fOU) and rough Heston (rHeston) processes. Empirical case studies include:
\begin{itemize}
	\item Estimating the Hurst parameter from historic realized volatility in financial markets;
	\item Quantifying long-range dependence in Li-ion battery degradation data \cite{boros2024deep}.
\end{itemize}
Through extensive numerical and statistical evaluation, we demonstrate that SigMA outperforms existing benchmarks—CNNs \cite{stone2020calibrating}, LSTMs \cite{boros2024deep}, deep signature models \cite{Lyons_2019_DeepSignature}, and vanilla Transformers \cite{vaswani_2017}—in terms of accuracy, robustness, and model compactness. Our findings support that path signatures, when integrated with attention-based architectures, offer a robust and scalable solution to parameter estimation in stochastic systems with rough or persistent temporal behavior.

The remainder of this paper is organized as follows. In Section \ref{sec-preliminaries}, we formulate the problem of interest and  give a brief introduction to the considered stochastic processes, path signature tools and self-attention mechanism that will be used in this paper. The specific architecture of the SigMA is clarified in Section \ref{sec-methodology}. 
In Section \ref{sec-experiments}, we begin with numerical examples to explore the optimal architecture of the SigMA model, then we conduct rich examples to verify its performance, comparing it with several well-known benchmark models. Finally, we conclude with Section \ref{sec-conclusion}.

\section{Problem setting and framework}
\label{sec-preliminaries}
In this section, we establish a framework for parameter estimation from observed sample paths of fBm-driven stochastic processes. We begin by formalizing the estimation task (learning problem) in Section \ref{subsec-preliminaries-formalization}. Subsequently, we introduce the relevant stochastic processes in Section \ref{subsec-preliminaries-rough} and review the core components of the approach: the self-attention mechanism and the path signature transform in Sections \ref{subsec-preliminaries-transformer} and \ref{subsec-preliminaries-signature}.

\subsection{Problem formulation}
\label{subsec-preliminaries-formalization}		
Let $\mathbf{X}^{(n)} := (X_0, X_1, \dots, X_{n-1}) \in \mathbb{R}^{n \times d}$ denote a discrete observation of a $d$–dimensional stochastic process, where $n \in \mathbb{N}$ is the sequence length and $d \in \mathbb{N}$ is the dimensionality of each observation. The law of the underlying continuous–time process depends on an unknown parameter vector $\boldsymbol{\theta}_{\mathcal{V}}\in\mathbb R^{d_{\mathcal{V}}}$ that we wish to estimate.

To this end, we introduce a parametric predictor model
\[
\mathcal{M}(\cdot\,; \boldsymbol{\theta}_{\mathcal{M}}) \colon \mathbb{R}^{n \times d} \to \mathbb{R}^{d_\mathcal{V}},
\]
with learnable model parameters $ \boldsymbol{\theta}_{\mathcal{M}}$. The model $\mathcal{M}$ takes the observed path $\mathbf{X}^{(n)}$ as input and outputs an estimate of $\boldsymbol{\theta}_{\mathcal{V}}$.

We adopt a supervised learning framework in which the model $\mathcal{M}$ is trained on a synthetic dataset $\{(\mathbf{X}^{(n)})^{(i)}, \boldsymbol{\theta}_{\mathcal{V}}^{(i)})\}_{i=1}^{N_{\text{train}}}$, where each path $(\mathbf{X}^{(n)})^{(i)}$ is generated from a known stochastic model with known parameter $\boldsymbol{\theta}_{\mathcal{V}}^{(i)}$. The model parameters $\boldsymbol{\theta}_{\mathcal{M}}$ are learned by minimizing a loss function $\mathcal{L}: \mathbb{R}^{d_{\mathcal{V}}} \times \mathbb{R}^{d_{\mathcal{V}}} \to \mathbb{R}_{+}$:
\[
\widehat{\boldsymbol{\theta}}_{\mathcal{M}} = \arg\min_{\boldsymbol{\theta}_{\mathcal{M}}} \; \frac{1}{M} \sum_{i=1}^{N_{\text{train}}} \mathcal{L}\big(\mathcal{M}((\mathbf{X}^{(n)})^{(i)}; 	\boldsymbol{\theta}_{\mathcal{M}}),\, \boldsymbol{\theta}_{\mathcal{V}}^{(i)}\big).
\]

Once trained, the learned map $\widehat{\mathcal M}:=\mathcal M(\cdot;	\widehat{\boldsymbol{\theta}}_{\mathcal{M}} )$ is subsequently deployed on unseen paths (observations), including real‐world data.  

\subsection{Stochastic processes}
\label{subsec-preliminaries-rough} 
In this section, we provide a brief formulation of the stochastic processes we study, namely the fractional Brownian motion (fBm), and   two fBm-driven models: the fractional Ornstein–Uhlenbeck (fOU), and the rough Heston (rHeston) processes.

We work on a fixed filtered probability space $(\Omega, \mathcal{F}, (\mathcal{F}_t)_{t \geq 0}, \mathbb{P})$ that supports all stochastic processes under consideration. In abstract form, we consider a class of stochastic differential equations (SDEs) of the form
\begin{equation}
	\mathrm{d}X(t) = \mu\big(X(t), \boldsymbol{\theta}_{\mathcal{V}}\big)\, \mathrm{d}t + \sigma\big(X(t), \boldsymbol{\theta}_{\mathcal{V}}\big)\, \mathrm{d}B^H(t), \quad X(0) = x_0 \in \mathbb{R}^d, \label{eq:SDE_fbm}
\end{equation}
where $X(t) \in \mathbb{R}^d$ is the state variable, $\boldsymbol{\theta}_{\mathcal{V}} \in \mathbb{R}^{d_{\mathcal{V}}}$ is the unknown parameter vector to be estimated, and $\{B^H_t\}_{t \ge 0}$ is a $d$-dimensional fBm with Hurst parameter $H \in (0,1)$. When $H = \frac{1}{2}$, $B^H$ reduces to the standard Brownian motion and \eqref{eq:SDE_fbm} becomes an Itô SDE. In general, however, $B^H$ is not a semimartingale unless $H = \frac{1}{2}$. 

We focus on three representative classes of fBm-driven processes:

\subsubsection{Fractional Brownian motion}
The simplest case is  $X(t) = B^H(t)$, with $\boldsymbol{\theta}_{\mathcal{V}} = H$. This process is centered Gaussian, $H$-self-similar, and has stationary increments, with a covariance function
\[
\mathbb{E}[B^H_s B^H_t] = \tfrac{1}{2}\big(s^{2H} + t^{2H} - |t - s|^{2H}\big), \quad s,t \geq 0.
\]

For $H > \tfrac{1}{2}$, the increments exhibit long-range dependence; for $H < \tfrac{1}{2}$, they exhibit rough anti-persistence. Almost surely, the sample paths are Hölder continuous of any order $\gamma < H$ \cite{biagini2008stochastic, nualart2006malliavin}.

There are  different integral representations of fBm, in particular  the characterisation provided by \cite{mandelbrot_1968_fbm} as
\begin{eqnarray}
	\label{eq-fBm1}
	B_t^H
	&=& \frac{1}{\Gamma\(\frac{1}{2}+H\)}\int_{-\infty}^{0} ((t-s)^{H-\frac{1}{2}}-(-s)^{H-\frac{1}{2}})dB_s \nno\\
	&+& \frac{1}{\Gamma\(\frac{1}{2}+H\)}\int_{0}^{t} (t-s)^{H-\frac{1}{2}}dB_s,\quad t\in[0,T]
\end{eqnarray}
where $\{B_t\}_{t\ge 0}$ is a standard Brownian motion and $\Gamma(\cdot)$ is the standard Gamma function. 

\subsubsection{Fractional Ornstein–Uhlenbeck process} 
The fOU process \cite{kleptsyna2002statistical,cheridito2003fractional} is the  fractional analogue of the OU process, where  the process,  $\{X^{fOU}_t\}_{t \geq 0}$,  solves
\[
dX^{fOU}_t = -\alpha (X^{fOU}_t- \mu) \, dt + \sigma \, dB^H_t, \quad X^{fOU}_0 = x_0,
\]
where $\alpha > 0$, $\sigma > 0$,  $\mu \in \mathbb{R}$, and $\{B^H_t\}_{t\ge 0}$ is an fBm with Hurst index $H \in (0,1)$. 
The integral solution is given  by
\[
X^{fOU}_t = \mu +e^{-\alpha t} (x_0 -\mu) + \sigma \int_0^t e^{-\alpha(t - s)} \, dB^H_s.
\]
In this case, the parameter vector to estimate  in Section \ref{subsec-preliminaries-formalization}  is $\boldsymbol{\theta}_{\mathcal{V}} = (H, \alpha, \mu, \sigma)$.

\subsubsection{Rough Heston process} 
The rHeston model \cite{el2019characteristic}, a non-Markovian extension of the classical Heston model, is another popular rough stochastic volatility model. It is a mean-reverting process that effectively captures both the volatility skew and volatility smiles observed in financial markets. In this model, the variance process, $\{X_t^{\mathrm{rH}}\}_{t \geq 0}$, satisifies 
\begin{eqnarray}
	\label{eq-rHeston}
	X_t^{rH} &=& X_0^{rH} + \frac{1}{\Gamma\(\frac{1}{2}+H\)}\int_0^t(t-s)^{H-\frac{1}{2}} \kappa_1 (\theta-X_s^{rH})\,ds \nno\\
	&+& \frac{1}{\Gamma\(\frac{1}{2}+H\)}\int_0^t\(t-s\)^{H-\frac{1}{2}} \kappa_2 \sqrt{X_s^{rH}}\,dB_s,\quad t\in[0,T],
\end{eqnarray}
where $H\in(0,\,\frac{1}{2})$, $X_0^{rH}\in\mathbb{R}^+$ is the initial value and $\kappa_1,\,\kappa_2,\,\theta>0$.

Equivalently, defining the fractional kernel $K_H(t) := t^{H - \frac{1}{2}} / \Gamma(H + \frac{1}{2})$, we may rewrite \eqref{eq-rHeston} as
\[
X_t^{\mathrm{rH}} = X_0^{rH}  + \kappa_1 \int_0^t K_H(t - s)(\theta - X_s^{\mathrm{rH}})\,ds + \kappa_2 \int_0^t K_H(t - s) \sqrt{X_s^{\mathrm{rH}}} \, dB_s.
\]
In this case, the parameter vector to estimate  in Section \ref{subsec-preliminaries-formalization}  is $\boldsymbol{\theta}_{\mathcal{V}} =(H, \kappa_1, \kappa_2, \theta)$.			

\subsection{Self-attention mechanism}
\label{subsec-preliminaries-transformer}						
Self-attention is a core component of Transformer architectures \cite{vaswani_2017}  and provides a mechanism for modeling dependencies between elements in a sequence, regardless of their relative positions. Formally, given an input sequence $\mathbf{X}^{(n)} = (X_0, \ldots, X_{n-1}) \in \mathbb{R}^{n \times d}$, the attention mechanism computes a new representation for each element based on a weighted aggregation of the entire sequence.

We define the self-attention map as a function
\[
\mathcal{A} : \mathbb{R}^{n \times d} \to \mathbb{R}^{n \times d_{\text{att}}},
\]
parametrized by three learnable projection matrices $W_Q, W_K, W_V \in \mathbb{R}^{ d \times d_{\text{att}}  }$ (typically $ d>d_\text{att} \in\mathbb{N}^+ $), and the self-attention map is then defined as
\begin{align}
	\label{eq-single-attention}
	\mathcal{A} (\mathbf{X}^{(n)}; \;\,W_Q,\,W_K,\,W_V)&:=\text{softmax}\left(\frac{QK^T}{\sqrt{d_{\text{att}}}}\right)V \in \mathbb{R}^{n \times d_{\text{att}}}
\end{align}
where $Q := \mathbf{X}^{(n)} W_Q,  \; K := \mathbf{X}^{(n)} W_K, \; V := \mathbf{X}^{(n)} W_V$ are    the query, key, and value matrices,  and the softmax in \eqref{eq-single-attention} is applied row-wise.  Each output row is a weighted sum of all value vectors, where weights are determined by the similarity between the query and key vectors.\footnote{Each input vector $x_i \in \mathbb{R}^d$ is projected into three latent vectors: 		 $	q_i := W_Q x_i, \; k_i := W_K x_i, \; v_i := W_V x_i$.		  	The attention output corresponding to $x_i$ is given by
	\begin{equation*}
		\mathrm{Attn}(x_i) = \sum_{j=1}^n \alpha_{ij} v_j, \quad \text{with} \quad \alpha_{ij} = \frac{\exp\left( \langle q_i, k_j \rangle / \sqrt{d_{\text{att}}} \right)}{\sum_{j'=1}^n \exp\left( \langle q_i, k_{j'} \rangle / \sqrt{d_{\text{att}}} \right)},
	\end{equation*}
	where $\langle \cdot, \cdot \rangle$ denotes the standard Euclidean inner product, and $\alpha_{ij}$ are the attention weights obtained via a softmax normalization.}

Instead of performing a single self-attention function, it is  often beneficial to apply a multi-head self-attention function,  which performs self-attention functions in parallel  to the different components of  $\mathbf{X}^{(n)}$  and then linearly projects the concatenated outputs into an appropriate space.

Specifically, for $h\in\mathbb{N}^+$ parallel heads, we introduce distinct projection matrices for each head $i \in \{1, \dots, h\}$:
\[
Q_{i} := \mathbf{X}^{(n)}_i W_{Q_i}, \quad K_{i} := \mathbf{X}^{(n)}_i  W_{K_i}, \quad V_i := \mathbf{X}^{(n)}_i  W_{V_i},
\]
with $\mathbf{X}^{(n)}_i$ denotes the input to head $i$, obtained via linear projection and $W_{Q_i}, W_{K_i}, W_{V_i} \in \mathbb{R}^{d \times d_{\text{att},i}}$. 

Each head computes its own attention output (using  Equation~\eqref{eq-single-attention}):
\[
\text{head}_i = \mathcal{A}(\mathbf{X}_i^{(n)}; W_{Q_i}, W_{K_i}, W_{V_i}) \in \mathbb{R}^{n \times d_{\text{att},i}},
\]
and the outputs of all heads are concatenated and projected back to the original feature dimension via a learned matrix $W_O \in \mathbb{R}^{h \; d_{\text{att}_i} \times d}$, which results in the multi-head self-attention function, defined as
\begin{eqnarray}
	\label{eq-multi-attention}
	\text{Multi-Head}(\mathbf{X}^{(n)};\,W^{Q_1},\,\ldots,\,W^{Q_h},\,W^{K_1},\,\ldots,\,W^{K_h},\,W^{V_1},\,\ldots,\,W^{V_h},\,W^O) \nno\\
	:= \text{Concat}(\text{head}_1,\,\ldots,\,\text{head}_h)W^O,
\end{eqnarray}

Multi-head self-attention mechanisms allow the model to jointly attend to information from different representation subspaces at different positions whereas a single-head mechanism restricts the model to one sub-space.

In the architecture, self-attention is applied to path features encoded by the signature transform (see Section~\ref{subsec-preliminaries-signature}), acting as a non-local feature aggregator over time.

\subsection{Path signatures}
\label{subsec-preliminaries-signature}
A path signature transform \cite{Lyons_1998_decay} is a feature extraction technique for sequential data, particularly effective for multivariate time series analysis. 
It characterizes a path $X=\{x_t | x_t=(x_t^1,\ldots,x_t^d)\in\mathbb{R}^d,\,t\in[a,\,b]\}$, $a,\,b\in\mathbb{R}$ through a sequence of iterated integrals, capturing complex temporal dependencies and interactions between different dimensions of the data. 
Formally, the signature in the Stratonovich sense is defined as follows
\begin{eqnarray}
	\label{eq-signature}
	&&\text{Sig}(X):=\left(1,\,\text{Sig}_1(X),\,\text{Sig}_2(X),\,\ldots,\,\text{Sig}_i(X),\,\ldots\right),\nno\\
	&&\text{Sig}_i(X) = \underset{a<t_1<\cdots<t_i<b}{\int\cdots\int}\circ dx_{t_1}\otimes\cdots\otimes\circ dx_{t_i},\quad i=1,\,2,\,\ldots,
\end{eqnarray}
where $\circ$ denotes the Stratonovich integral and $\otimes$ denotes the tensor product. 
Here, the "zeroth" term $\text{Sig}_0(X)$, by convention, is set to 1. 
Moreover, we denote the truncated signature $\text{Sig}^N(X)=\left(1,\,\text{Sig}^N_1(X),\,\ldots,\,\text{Sig}^N_N(X)\right)$ with the truncation order $N\in\mathbb{N}^+$. 

We summarize two properties of signatures that make them attractive in learning from paths and we present these here without any proof. 
\begin{proposition}[Uniqueness (\cite{Lyons_2010_uniqueness})]
	\label{prop-uniqueness}
	Let $\mathbf{X}^{(n)}={(X_0,\,\ldots,\,X_{n-1})}^T\in\mathbb{R}^{n\times d}$ be a path stream, and define its time-augmented version as 
	$$\mathbf{\widehat{X}}^{(n)}:={((X_0,\,t_0),\,\ldots,\,(X_{n-1},\,t_{n-1}))}^T\in\mathbb{R}^{n\times(d+1)}.$$
	Then the signature $\text{Sig}(\mathbf{\widehat{X}}^{(n)})$ can uniquely determine $\mathbf{X}^{(n)}$ up to a translation.
\end{proposition}
Proposition~\ref{prop-uniqueness} implies that there is no information lost in the signature transform. 
The time-augmented path is necessary because the signature transform is invariant to time reparameterizations, meaning it encodes only the data arrival order, not the precise timing. 
\begin{proposition}[Factorial decay (\cite{Lyons_1998_decay})]
	\label{prop-decay}
	There exists a constant $C(\mathbf{X}^{(n)})$ depending on the path stream $\mathbf{X}^{(n)}$ such that $$\left\|\text{Sig}_i(\mathbf{X}^{(n)})\right\| \leq \frac{C(\mathbf{X}^{(n)})^i}{i!},$$ where $\left\|\cdot\right\|$ is any tensor norm on ${(\mathbb{R}^d)}^{\otimes i}$.
\end{proposition}
Proposition~\ref{prop-decay} indicates that the terms of the signature decay factorially in size. This ensures that the most significant contributions are given by the lower order terms which justifies the use of truncated signatures.
\begin{proposition}[Chen's identity (\cite{chen1958integration})]
	\label{prop-chen}
	Let $X:[s,t]\to\mathbb{R}^d$ be a path of bounded variation, and let $u\in(s,t)$ be an intermediate time. 
	The signature of the entire path $\text{Sig}(X_{[s, t]})$ satisfies
	$$\text{Sig}(X_{[s, t]})=\text{Sig}(X_{[s,u]})\otimes\text{Sig}(X_{[u,t]}).$$
\end{proposition}
Proposition~\ref{prop-chen} shows that the signature of a long trajectory implicitly contains the signatures of its subpaths.
In practice, this algebraic decomposition of path signatures justifies the "lifting" procedure within SigMA, as detailed in Section~\ref{subsec-methodology-lifting}.

\section{Methodology}
\label{sec-methodology}

\begin{figure}[h]	
	\centering
	\includegraphics[width=0.8\textwidth]{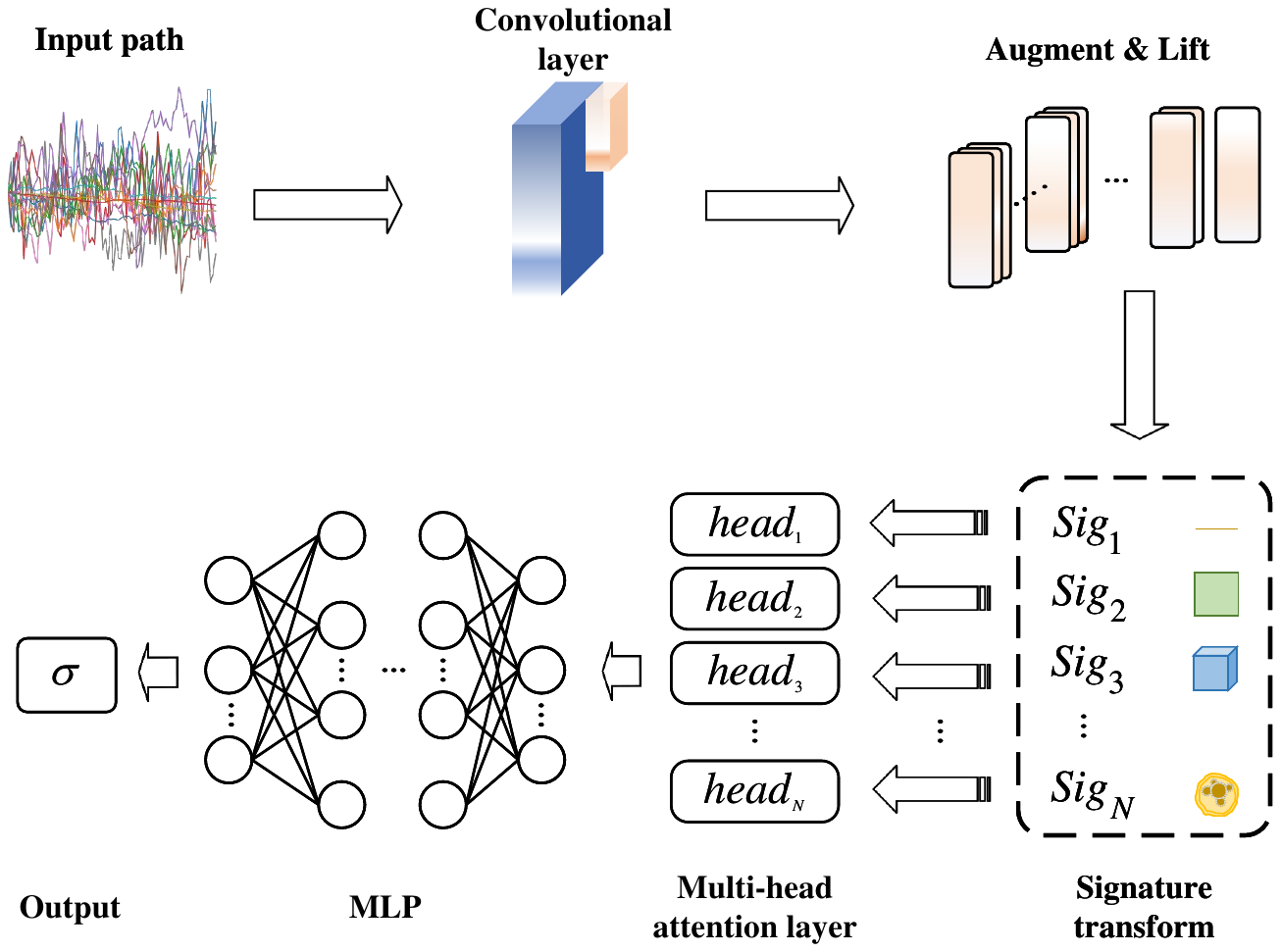}	
	\caption{Architecture of the SigMA model. }
	\label{fig-structure-SigMA}
\end{figure}

In this section, we describe the architecture of the SigMA model, as shown in Figure \ref{fig-structure-SigMA}; the rationale behind these design choices is discussed in Section \ref{subsec-experiments-sensitivity}.

\subsection{Convolutional feature extraction}
\label{subsec-methodology-convolutional}
Consider a data stream  $\mathbf{X}^{(n)}={(X_0,\,\ldots,\,X_{n-1})}^T\in\mathbb{R}^{n\times d}$, interpreted as a discretized path. It is beneficial to apply a feature mapping before computing the signature transform (see \cite{Lyons_2019_DeepSignature}). Therefore, at the input of SigMA, a convolutional layer  $\Phi(\mathbf{X}^{(n)};\,W^{\Phi},B^{\Phi}):\mathbb{R}^{n\times d}\mapsto\mathbb{R}^{\widetilde{n}\times\widetilde{d}}$ with kernel size $k\in\mathbb{N}^+$ and stride $s\in\mathbb{N}^+$ is applied to perform the feature mapping
\begin{equation}
	\label{eq-convolutional-layer}
	\Phi(\mathbf{X}^{(n)};\,W^{\Phi},B^{\Phi})=W^{\Phi}\star \mathbf{X}^{(n)} + \mathbf{1}_{\widetilde{n}}B^{\Phi},
\end{equation}
with
\begin{equation}
	\label{eq-cross-correlation}
	(W^{\Phi}\star \mathbf{X}^{(n)})_{i,j}=\sum\limits_{i_1=1}^{k}\sum\limits_{i_2=1}^{d}W^{\Phi}_{j,i_1,i_2}\mathbf{X}^{(n)}_{s(i-1)+i_1,i_2},\quad i=1,\,\ldots,\,\widetilde{n},\,j=1,\,\ldots,\,\widetilde{d},
\end{equation}
where $W^{\Phi}\in\mathbb{R}^{\widetilde{d}\times k\times d}$ and $B^{\Phi}\in\mathbb{R}^{\widetilde{d}}$ are the weights and biases of this convolutional layer, respectively, while $\widetilde{n}\in\mathbb{N}^+$ and $\widetilde{d}\in\mathbb{N}^+$ represent the dimensions of outputs. 
We denote $\star$ as the valid cross-correlation operator and $\mathbf{1}_{\widetilde{n}}\in\mathbb{R}^{\widetilde{n}}$ is an all-ones column vector. This convolutional layer is shown in Section \ref{subsubsec-sensitivity-structure} to improve the accuracy of the SigMA model.				After the feature mapping, according to Proposition~\ref{prop-uniqueness}, it is beneficial to augment the output with time before taking the signature transform. 
This augmentation enables the signature to encode the exact arrival times and the specific $\Delta t$ between observations. 
Consequently, SigMA can natively process irregularly sampled data and adapt to varying sampling frequencies. 

\begin{remark}
	\label{remark-convolutional-layer}
	The convolutional layer in SigMA serves not only as a feature encoder but also enables SigMA to handle high-dimensional processes. 
	For a path of dimension $d$, the number of terms in its signature truncated at order $N$ grows exponentially as $O(d^N)$. 
	If the signature were computed directly on high-dimensional raw data, the computational cost would become expensive. 
	SigMA mitigates this challenge by first projecting the raw path $\mathbf{X}^{(n)}\in\mathbb{R}^{n\times d}$ into a lower-dimensional latent representation of size $\tilde{d}$ before computing signatures. 
	Therefore, the subsequent signature transform is independent of the raw input dimension, and the size of the truncated signature scales as $O(\tilde{d}^N)$. 
	This decouples the exponential growth of the signature tensor from the original input dimension, enabling SigMA to scale efficiently to higher-dimensional SDEs without causing a parameter explosion in downstream layers. 
\end{remark}

\subsection{Signature transform on lifted paths}
\label{subsec-methodology-lifting}
Taking the signature transform can extract the geometric features from original paths while filtering out irrelevant information, however, it will consume the stream-like nature of the input data (see \cite{Lyons_2019_DeepSignature}).
Let us consider a stream of data $\mathbf{Y}^{(\widetilde{n})}={(Y_0,\,\ldots,\,Y_{\widetilde{n}-1})}^T\in\mathbb{R}^{\widetilde{n}\times\widetilde{d}}$, which is the output of the convolutional layer and has been augmented. 
Here, $\widetilde{d}$ denotes the number of features, $\widetilde{n}$ denotes the length of the sequence and is assumed to be even.  

We lift the data with the stride equals to half of the sequence length before taking the signature transform as follows
\begin{equation}
	\label{eq-lifting}
	l(\mathbf{Y}^{(\widetilde{n})}):=(\mathbf{Y}^{(\widetilde{n})}_{0:\frac{\widetilde{n}}{2}},\,\mathbf{Y}^{(\widetilde{n})}_{0:\widetilde{n}}),
\end{equation}
where $\mathbf{Y}^{(\widetilde{n})}_{0:i}={(Y_0,\,\ldots,\,Y_{i-1})}^T\in\mathbb{R}^{i\times\widetilde{d}}$ for $i=\frac{\widetilde{n}}{2},\,\widetilde{n}$.  Here, the stride controls the overlap between successive lifted segments, and each element in the lifted data $l(\mathbf{Y}^{(\widetilde{n})})$ is sequential data. 
Then the output of taking the signature transform with this lifted data becomes
\begin{equation}
	\label{eq-lifting-signature}
	\text{Sig}^N(l(\mathbf{Y}^{(\widetilde{n})}))={(\text{Sig}^N(\mathbf{Y}^{(\widetilde{n})}_{0:\frac{\widetilde{n}}{2}}),\,\text{Sig}^N(\mathbf{Y}^{(\widetilde{n})}_{0:\widetilde{n}}))}^T.
\end{equation}
It should be noted that setting the stride equal to half the sequence length in Equation \eqref{eq-lifting} represents a validated trade-off between model accuracy and complexity, as we have investigated the effect of stride values on both the accuracy and complexity of the neural network models in Section \ref{subsubsec-sensitivity-stride}. 

As shown in Equation~\eqref{eq-signature}, the path signature is defined as global iterated integration. 
However, when a truncated signature is computed over the entire trajectory, geometric information from different temporal regions becomes compressed into a single feature vector. 
By introducing lifted segments and computing signatures on these subsegments separately, the model receives explicit access to both local and global geometric features of the path. 
This provides a practical mechanism for constructing a multi-scale representation of the trajectory. 
Such segmented or streaming applications of signatures are commonly used in machine learning settings, where signatures are computed on successive windows of a time series in order to capture localized geometric information~\cite{chevyrev2025primer, Lyons_2019_DeepSignature}. 

\subsection{Multi-head self-attention on signature features}
\label{subsec-methodology-transformer}
Recall that the signature transform can be considered as a feature mapping which helps to extract the information from the input data while filtering out irrelevant information.   The signature transform yields a high-dimensional feature representation that typically requires additional processing. A common approach is to use a FNN, as in Deep Signature Networks \cite{Lyons_2019_DeepSignature}. In addition to an FNN, we incorporate a self-attention layer, which is particularly effective for joint multi-parameter estimation.	

Consider the output obtained by taking the signature transform as follows\footnote{The matrix stacks signature terms across truncation levels and lifted subpaths.}
\begin{eqnarray}
	\label{eq-signature-output}
	\text{Sig}^N(l(\mathbf{Y}^{(\widetilde{n})}))
	&=&{(\text{Sig}^N(\mathbf{Y}^{(\widetilde{n})}_{0:\frac{\widetilde{n}}{2}}),\,\text{Sig}^N(\mathbf{Y}^{(\widetilde{n})}_{0:\widetilde{n}}))}^T \nno\\
	&=&
	{\begin{pmatrix}
			\text{Sig}_1^N(\mathbf{Y}^{(\widetilde{n})}_{0:\frac{\widetilde{n}}{2}})&\cdots&\text{Sig}_N^N(\mathbf{Y}^{(\widetilde{n})}_{0:\frac{\widetilde{n}}{2}}) \\\
			\text{Sig}_1^N(\mathbf{Y}^{(\widetilde{n})}_{0:\widetilde{n}})&\cdots&\text{Sig}_N^N(\mathbf{Y}^{(\widetilde{n})}_{0:\widetilde{n}}) 
	\end{pmatrix}}_{2\times\frac{\widetilde{d}^{N+1}-\widetilde{d}}{\widetilde{d}-1}}.
\end{eqnarray}			    
We then apply a self-attention function, as shown in Equation~\eqref{eq-single-attention}, resulting in the output $\mathcal{A}(\text{Sig}^N(l(\mathbf{Y}^{(\widetilde{n})})))$, where we omit the weight matrices for simplicity.

As discussed in Section~\ref{subsec-preliminaries-transformer}, which is also verified in Section \ref{subsubsec-sensitivity-truncation}, using a multi-head attention mechanism is empirically beneficial instead of a single-head one. 
Therefore, we employ an $N$-head self-attention function, with each head corresponding to a distinct signature level, as follows
\begin{equation}
	\text{Multi-Head}(\text{Sig}^N(l(\mathbf{Y}^{(\widetilde{n})})))=\text{Concat}(\text{head}_1,\,\ldots,\,\text{head}_N)\widetilde{W}_O,
\end{equation}
where $\text{head}_i=\mathcal{A}({(\text{Sig}_i^N(\mathbf{Y}^{(\widetilde{n})}_{0:\frac{\widetilde{n}}{2}}),\,\text{Sig}_i^N(\mathbf{Y}^{(\widetilde{n})}_{0:\widetilde{n}}))}^T)$ for $i=1,\,\ldots,\,N$. 
Here we also omit all the weight matrix except the linear projection matrix $\widetilde{W}_O$ for the output. 

The design of the attention mechanism is motivated by structural properties of the signature transform.
Path signatures form a graded tensor algebra representation of trajectories, where different truncation levels correspond to geometric interactions of different orders~\cite{LyonsCaruanaLevy2007, FrizHairer2020}. 
Functions of path signatures are known to be universal approximators for functionals of streams~\cite{KidgerLyons2020}, and the expected signature uniquely characterizes the law of stochastic processes under mild conditions~\cite{ChevyrevOberhauser2018}. 
These results provide a theoretical justification for using signature features for parameter inference. 
In practice, truncated signatures produce a hierarchy of features with rapidly decaying magnitude across orders. 
Allowing the model to process these levels separately enables the architecture to assign different importance weights to different interaction orders. 
The multi-head attention layer therefore acts as an adaptive mechanism that learns which signature levels are most informative for the inference task.
While this mechanism is primarily motivated by these structural properties, its effectiveness is ultimately validated empirically in the experiments. 

After the $N$-head self-attention layer, we need to  project the output to an appropriate dimension using another multilayer perceptron (MLP) architecture, which has been shown to be important for the robustness of the SigMA model in Section \ref{subsubsec-sensitivity-structure}.

Now we can clarify the specific architecture of the SigMA as below: 
\begin{itemize}[noitemsep]
	\item a convolutional layer with 3 channels and kernel size 3;
	\item augmentation with time and original values;
	\item lifting performed as Equation~\eqref{eq-lifting} with stride equal to half of the sequence length;
	\item taking the signature transform with truncation order $N=3$;
	\item a multi-head self-attention layer with the number of heads $h=N=3$ and each different head corresponds to a different signature level;
	\item a MLP with 5 hidden layers, each of size 32 and ReLU activation;
	\item a non-linear transform with sigmoid function.
\end{itemize}

\section{Numerical and empirical experiments}
\label{sec-experiments}
In this section, we conduct experiments to verify the advantages of the architecture described in Section~\ref{sec-methodology} and motivate the architectural choices. 		To this end, we compare SigMA\footnote{Source code is available on \url{https://github.com/changanluoxue/SigMA.git}} with four other neural network-based models. 
The first model is a  CNN proposed by \cite{stone2020calibrating}.\footnote{Source code is available on \url{https://github.com/henrymstone/rough-calibration-using-CNNS}}
The second model is a signature-based (DeepSigNet) model proposed by \cite{Lyons_2019_DeepSignature}.\footnote{Source code is available on \url{https://github.com/patrick-kidger/Deep-Signature-Transforms}}
The third model is a modified Transformer based on \cite{vaswani_2017}. 
The fourth model is a LSTM model proposed by \cite{csanady2024parameter}.\footnote{Source code is available on \url{https://github.com/aielte-research/LMSParEst}}
The specific architectures of these four models are detailed in~\ref{apdix-architecture}. 

This section is divided into six subsections.  Firstly, we clarify the synthetic data generation and some implementation details. 
In Section \ref{subsubsec-sensitivity-implementation}, we perform sensitivity analysis to determine the optimal architecture of the SigMA model. 
Next, we discuss the single and multiple parameters estimation problems in Sections~\ref{subsec-experiments-single} and ~\ref{subsec-experiments-multiple}. 
In Section~\ref{subsec-experiments-empirical}, we present two empirical studies: one using realized volatility data from financial markets and another employing Li-ion battery degradation data. 
Finally, based on the presented results, we summarize the differences between SigMA and comparable models in terms of architectural designs, applicable scenarios, and performance in Section~\ref{subsec-experiments-summary}. 

\subsection{Data generation and implementation details}
\label{subsubsec-sensitivity-implementation}
Before conducting the experiments, we detail the generation of the synthetic paths used in the study and provide implementation specifics to ensure full reproducibility of the results. 
To generate datasets containing paths with varying characteristics, we first sample five Hurst parameters, $H$, from two different probability distributions: the Uniform distribution\footnote{The set of possible $H$ values for fBm is $\{0.14, 0.23, 0.56, 0.67, 0.89\}$, whereas the fOU employs another set of $\{0.55, 0.66, 0.72, 0.85, 0.94\}$.} on  $(0,1)$ and $(0.5,1)$ and the Beta $(1,9)$  distribution.\footnote{The set of possible $H$ values for rHeston is $\{0.04, 0.08, 0.11, 0.16, 0.21\}$, consistent with its theoretical construction in the rough regime.} 
Paths with $H$ from the Beta distribution are rougher than those from the Uniform distribution, while extending the Uniform's support to $H>\frac{1}{2}$ induces long memory. 
Using these different $H$ values, we simulate fBm, fOU, and rHeston paths with varying numbers of time steps $\{100, 500, 1000, 1500\}$, with each training/test data set containing 3000/1000 samples. 

We report the number of trainable parameters for all neural architectures as the hardware-independent proxy for computational complexity and memory footprint. 
We use the Cholesky decomposition to simulate fBm and subsequently generate fOU paths with $\alpha = 0.5$, $\mu = 0.15$, $\sigma = 0.2$, and $X_0^{fOU}=0.01$.  
For simulating rHeston paths, we apply the fast algorithm proposed by \cite{ma_2022_fast}, using the parameters $\kappa_1 = 0.1$, $\kappa_2 = 0.03$, $\theta = 0.3$, and $X_0^{rH}=0.01$.

We use the adaptive moment estimation (Adam) optimizer to train the different neural network models, with a batch size of 60, trained for  $150$ epochs, and a learning rate of $10^{-4}$. The root mean squared error (RMSE) is used as a loss function. 
All experiments were run on a system with CUDA 12.8, utilizing GPU acceleration. The specific versions of the key libraries were: PyTorch 2.8.0, Python 3.10.12.

The full implementation, data-generation scripts, and examples are available at \href{https://github.com/changanluoxue/SigMA.git}{[GitHub link]}.

\subsection{Sensitivity analysis of the SigMA architecture}
\label{subsec-experiments-sensitivity}
We first perform a sensitivity analysis on the stride and signature truncation order to identify the optimal configuration of the SigMA model, balancing accuracy with computational complexity. 
We then present illustrative examples to validate the effectiveness of both the convolutional layer and MLP components within the model.

\subsubsection{Sensitivity to the stride parameter}
\label{subsubsec-sensitivity-stride}
As discussed in Section \ref{subsec-methodology-lifting}, the lifting operation before taking the signature transform can preserve the stream-like nature of the data and then improve the performance of the SigMA model. 
The lifting operation with a smaller stride increases the frequency of the signature transform, which is time-consuming, and also extends the input length of the subsequent multi-head self-attention layer, thereby increasing the number of model parameters. 
Since the lifting operation increases the complexity of the SigMA model, it is crucial to determine an optimal stride that balances the model complexity with performance, which is the aim of this section. 

In this example, we use the simulated paths from Section \ref{subsubsec-sensitivity-implementation} to explore the optimal stride for paths of varying lengths. 
The stride of the lifting operation is adjusted from $1$ to $\frac{n}{2}$ (e.g., half the input sequence length), while other settings remain consistent with Section \ref{subsubsec-sensitivity-implementation}. 
It should be noted that $\textit{stride}=\textit{input length}$ effectively means that no lifting is applied, thus the maximum possible stride for a path stream is half its length. 

We present the test RMSEs for estimating the Hurst parameter across different stochastic processes using the SigMA model with varying stride in Table \ref{tab-sensitivity-stride}. 
From the results, two conclusions can be drawn. 
Firstly, the complex stochastic processes (i.e. the fOU and rHeston) tend to use small strides to achieve high accuracy. 
This observed sensitivity likely stems from the drift term in the fOU and rHeston, which the fBm does not include. 
Secondly, decreasing the stride significantly increases the number of parameters of the NN with negligible benefit in terms of accuracy for all models.
This suggests that a larger stride ($\textit{stride}=\frac{n}{2}$) should be chosen to strike a balance, as the slight improvement in model performance comes with a sharp increase in complexity.  

\begin{table}[H]
	\centering
	\caption{Test RMSEs for Hurst estimation across stochastic processes of varying lengths (100-1500), obtained using the SigMA model with strides varying from $1$ to $\frac{n}{2}$, are averaged over 3 training runs. Here, $H\sim\text{Uniform}(0,\,1)$ for the fBm, $H\sim\text{Uniform}(0.5,\,1)$ for the fOU, and $H\sim\text{Beta}(1,\,9)$ for the rHeston. The best accuracies across different strides are in bold.}
	\label{tab-sensitivity-stride}
	\smallskip
	\centering
	\renewcommand\arraystretch{1.1}
	\setlength{\tabcolsep}{6pt}
	\begin{tabular}{l c c c c c}
		\toprule
		\multirow{2}{*}{Input lengths}&\multirow{2}{*}{Strides}&\multicolumn{3}{c}{Stochastic processes}&\multirow{2}{*}{$\#$Params}\\
		\cline{3-5}
		&&fBm&fOU&rHeston&\\
		\hline
		\multirow{3}{*}{100}
		&1&1.94e-2&\textbf{2.51e-2}&\textbf{4.92e-2}&573306\\
		&10&\textbf{1.49e-2}&2.73e-2&5.07e-2&126906\\
		&50&1.58e-2&2.67e-2&5.25e-2&87226\\
		\hline
		\multirow{2}{*}{500}
		&50&8.71e-3&\textbf{1.72e-2}&\textbf{1.06e-2}&126906\\
		&250&\textbf{8.45e-3}&1.87e-2&1.09e-2&87226\\
		\hline
		\multirow{2}{*}{1000}
		&50&1.08e-2&2.37e-2&8.13e-3&176506\\
		&500&\textbf{6.05e-3}&\textbf{1.72e-2}&\textbf{7.48e-3}&87226\\
		\hline
		\multirow{2}{*}{1500}
		&50&\textbf{7.81e-3}&2.08e-2&\textbf{7.18e-3}&226106\\
		&750&2.71e-2&\textbf{2.05e-2}&7.61e-3&87226\\
		\bottomrule	
	\end{tabular}
\end{table}

\subsubsection{Effect of signature truncation order}
\label{subsubsec-sensitivity-truncation}
In this section, we aim to explore the effects of varying signature truncation orders on the accuracy of the SigMA model. 
In this experiment, we use the simulated paths from Section \ref{subsubsec-sensitivity-implementation} with lengths from $\{100,\,500\}$ and we vary the signature truncation order from $\{1,\,3,\,5\}$, while keeping other settings consistent with those in Section \ref{subsubsec-sensitivity-implementation}.
In addition to the SigMA model, we introduce a simpler version with a single-head self-attention layer, called SigSA (details in~\ref{apdix-architecture}). 
We also present results for the DeepSigNet model because it also incorporates the signature transform. 
Based on the finding from the previous section, we set a large stride as half of the input length. 

\begin{figure}[H]
	\centering
	\begin{subfigure}[b]{\textwidth}
		\centering
		\includegraphics[width=0.32\linewidth]{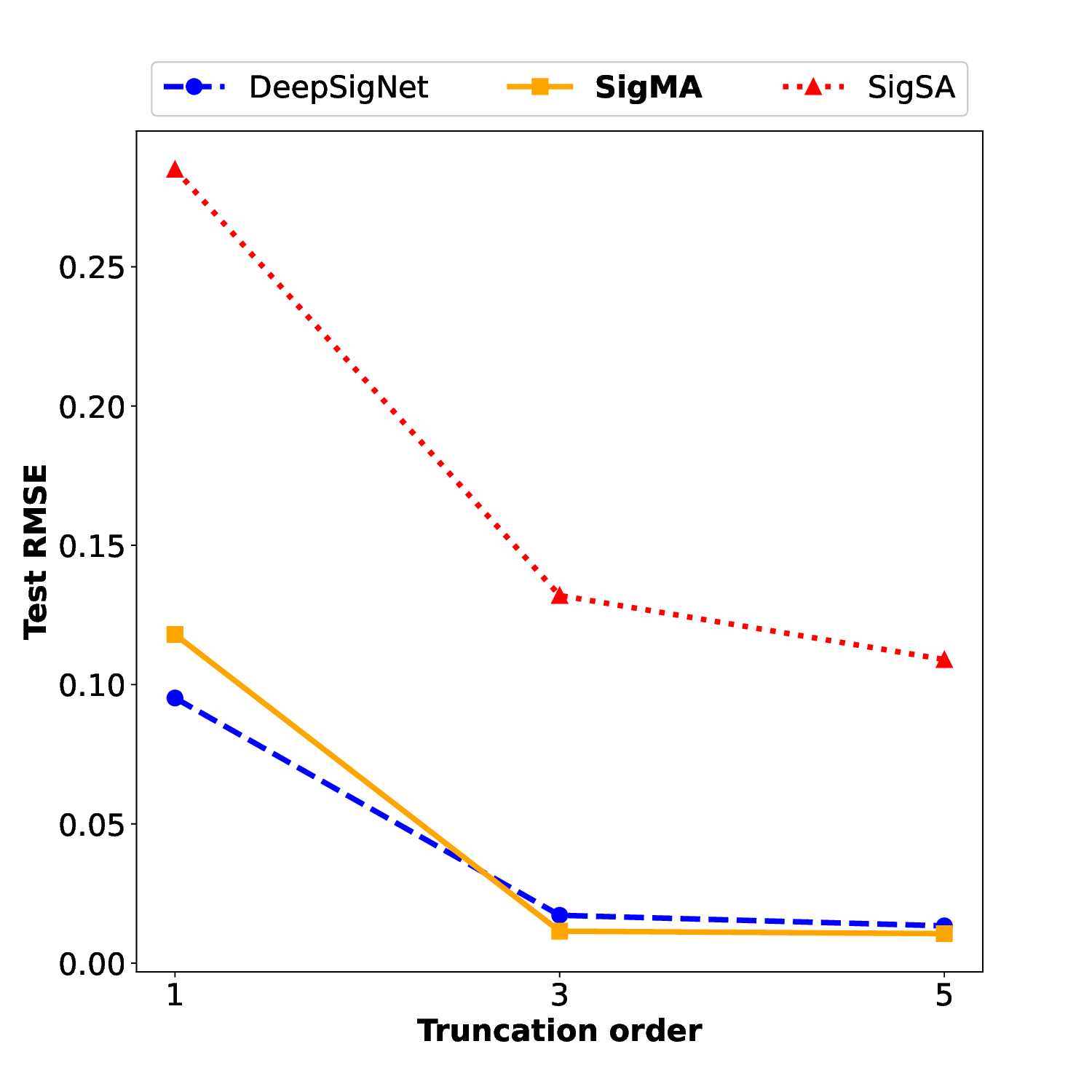}
		\includegraphics[width=0.32\linewidth]{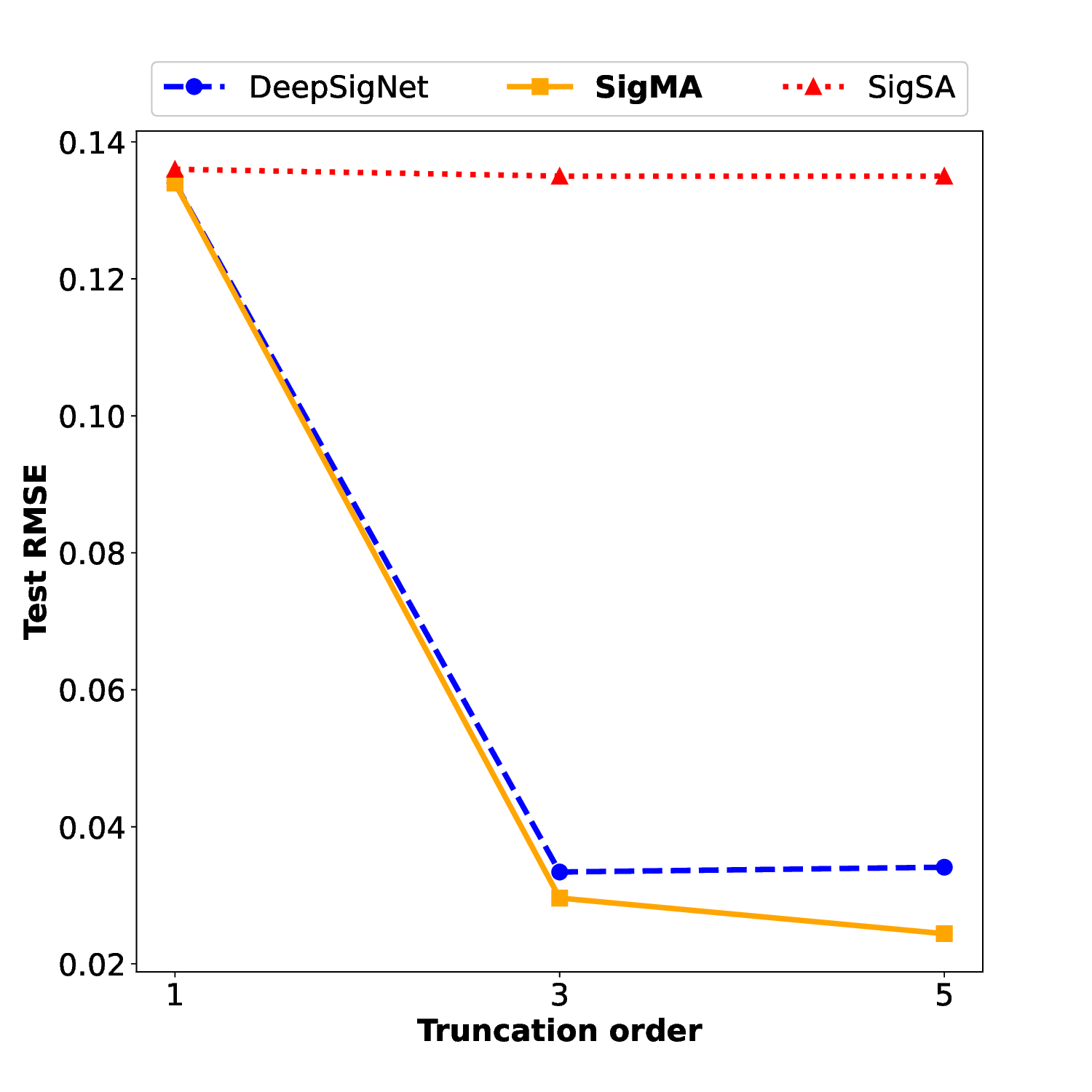}
		\includegraphics[width=0.32\linewidth]{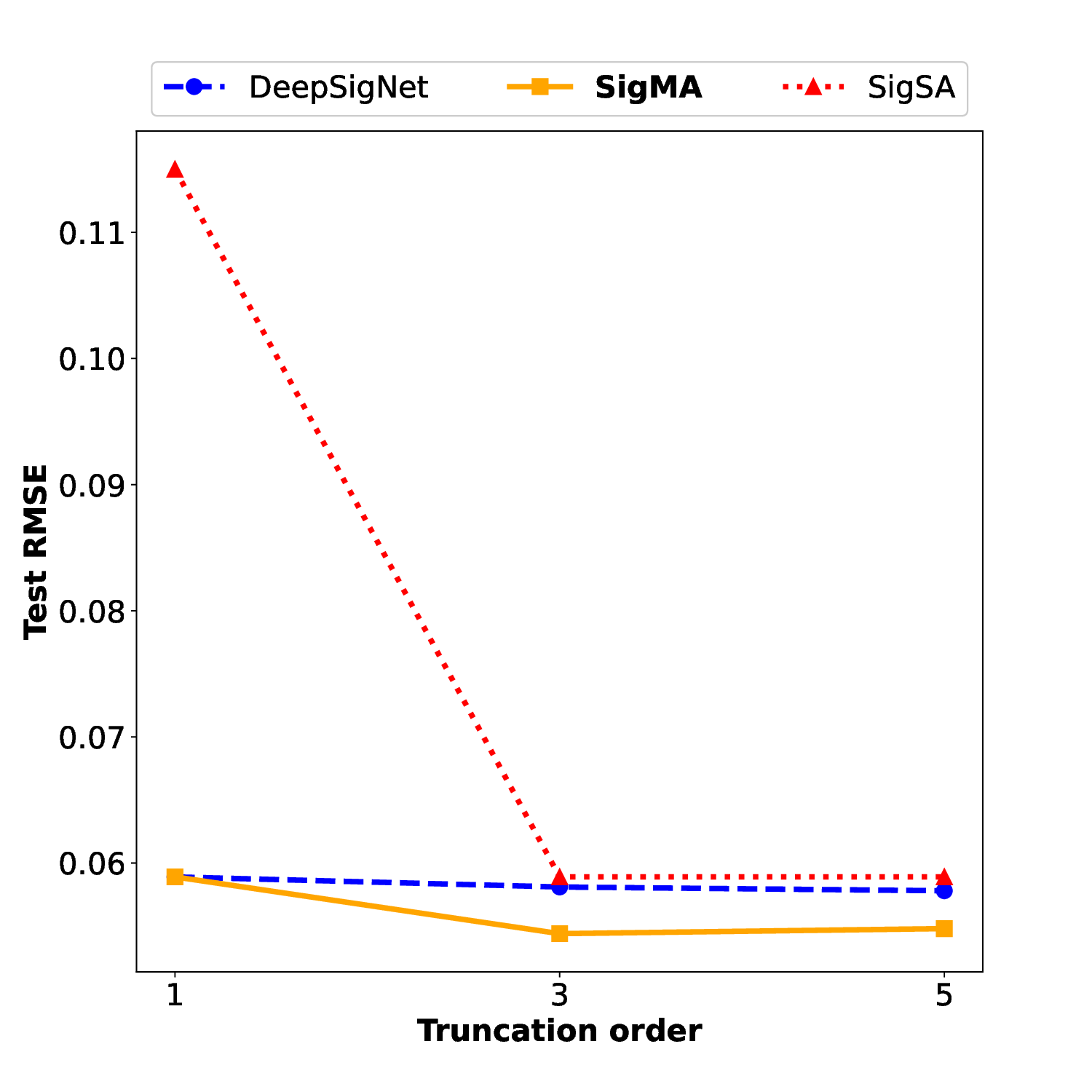}
		\caption{The input lengths equal to $100$}
	\end{subfigure}
	\begin{subfigure}[b]{\textwidth}
		\centering
		\includegraphics[width=0.32\linewidth]{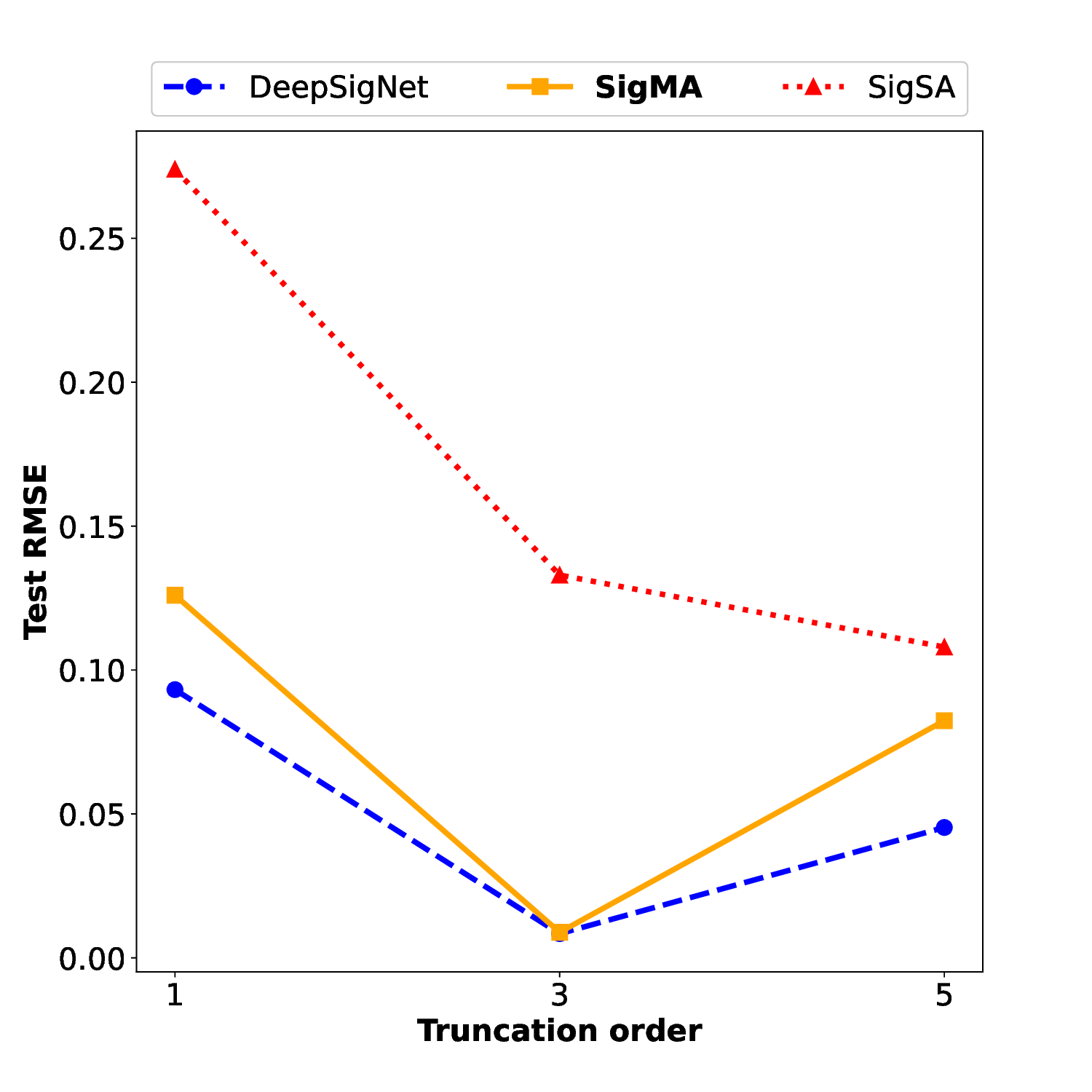}
		\includegraphics[width=0.32\linewidth]{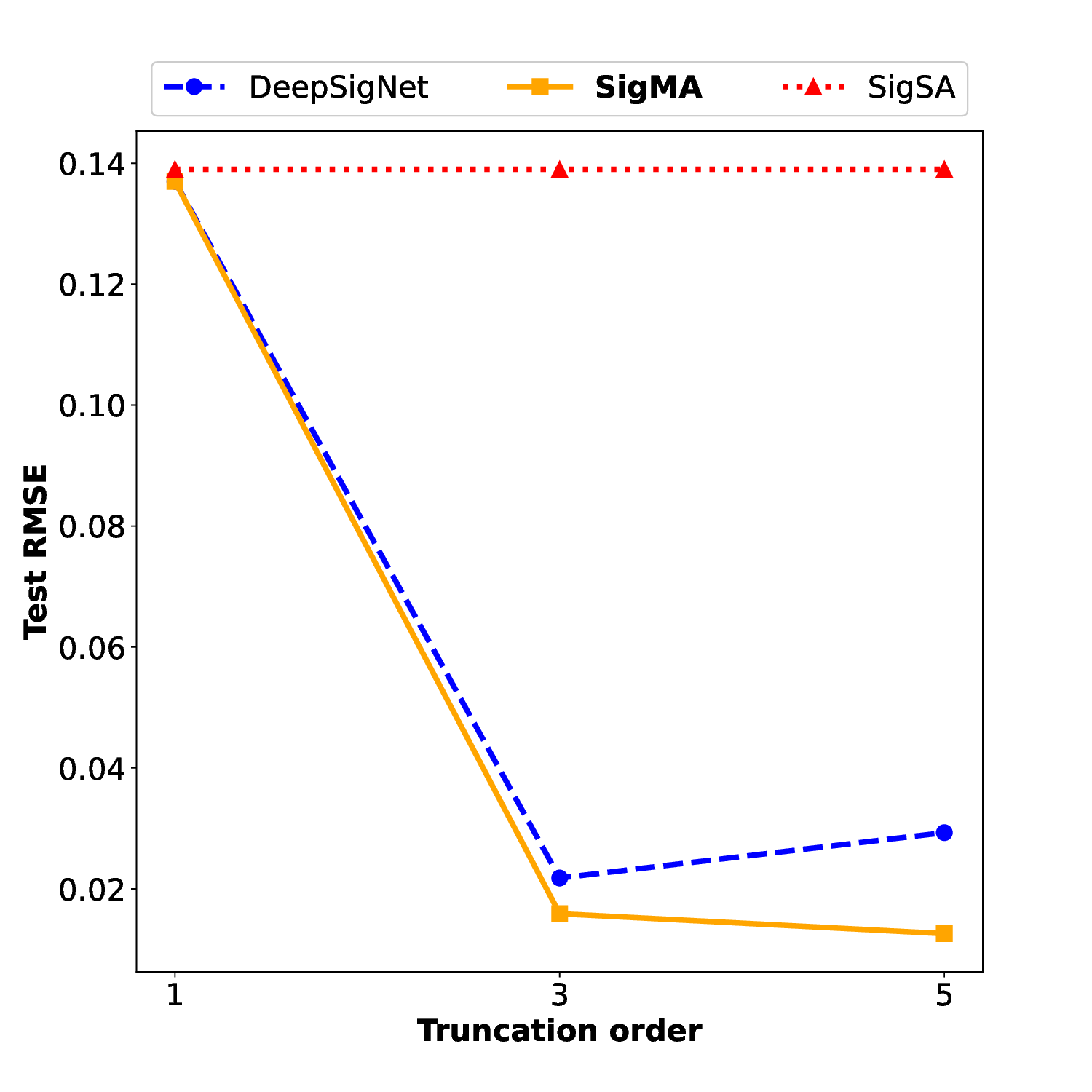}
		\includegraphics[width=0.32\linewidth]{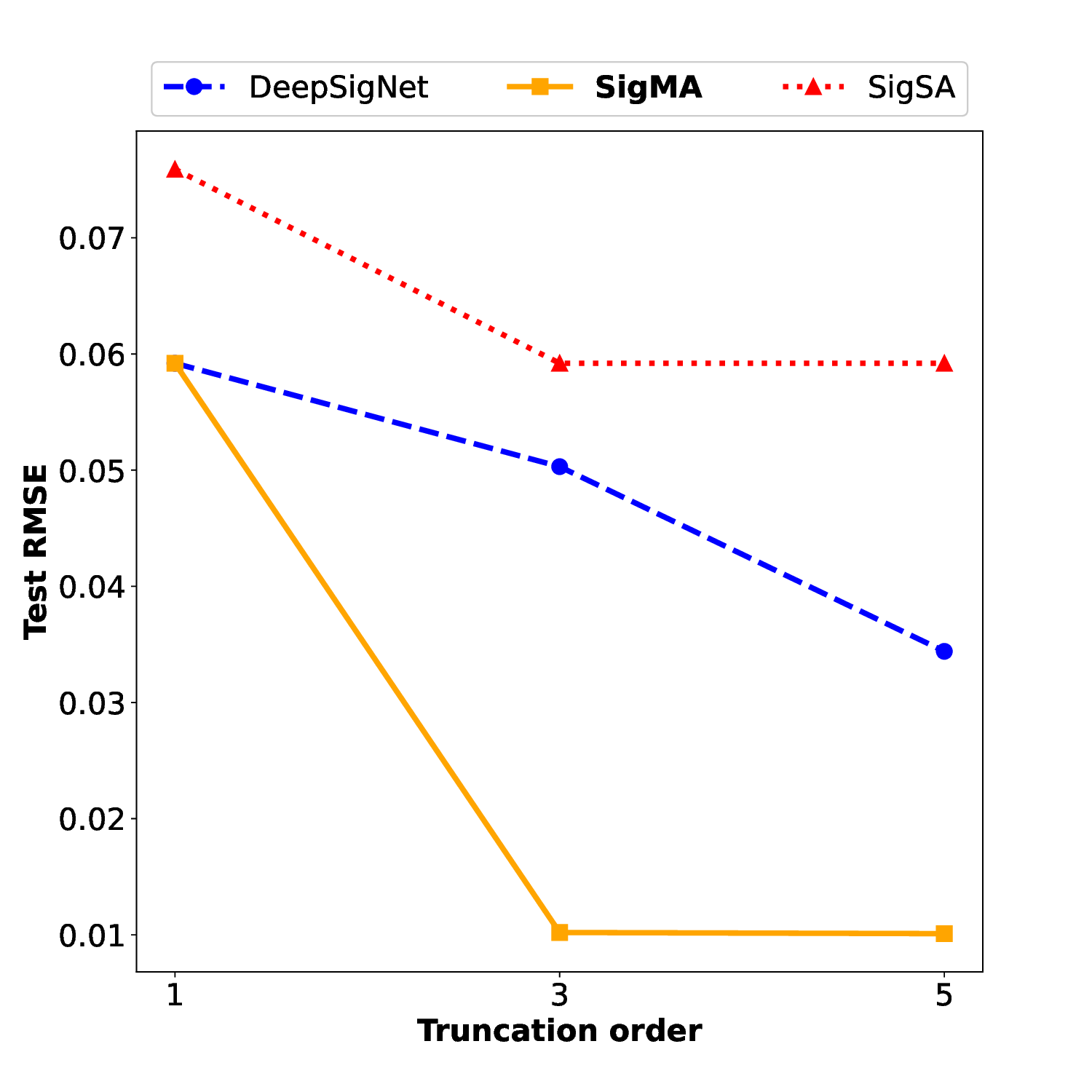}
		\caption{The input lengths equal to $500$}
	\end{subfigure}
	\caption{Test RMSE trends for Hurst estimation across stochastic processes (Left: fBms; Middle: fOU; Right: rHeston) of varying lengths (100-500), obtained using the signature-based NN models with varying signature truncation orders (1-5), are averaged over 3 training runs. Here, $H\sim\text{Uniform}(0,\,1)$ for the fBm, $H\sim\text{Uniform}(0.5,\,1)$ for the fOU, and $H\sim\text{Beta}(1,\,9)$ for the rHeston.}
	\label{fig-sensitivity-truncation}
\end{figure}

The results are presented in Figure \ref{fig-sensitivity-truncation} with details in Table \ref{tab-sensitivity-truncation}.
From the results, there are three conclusions to be drawn. 
Firstly, there are benefits for accuracy to increase the signature truncation order, particularly at low orders ($1$ to $3$). 
However, when the truncation order is sufficiently high ($3$ to $5$), the accuracy gains become negligible and overfitting issues may emerge. The complex stochastic processes (i.e. the fOU and rHeston) require higher optimal truncation orders compared to the simpler one (i.e. the fBm), which explains why the overfitting issues tend to occur in the fBm case. 
Furthermore, increasing the truncation order significantly increases the model complexity.
Secondly, a higher truncation order does not improve the performance of the SigSA model for the fOU paths. We attribute this to the single-head self-attention layer in the SigSA model, which is less effective at extracting information from the path signature compared to the multi-head self-attention layer.  
Thirdly, signature-based models with multi-head self-attention layers exhibit similar performance when the signature truncation order is set to 1, since the underlying stochastic processes are path-dependent and cannot be adequately represented by first-order increments alone. 
Notably, under such configurations, these models may even outperform the SigSA model, which is attributed to the limited capacity of a single-head self-attention mechanism to capture intricate path-dependent structures.
This suggests that we should select a signature truncation order of $3$ for the SigMA and DeepSigNet models to balance the accuracy and complexity. 

\begin{table}[H]
	\centering
	\rotatebox{90}{
	\begin{minipage}{\textheight}
	\caption{Test RMSEs for Hurst estimation across stochastic processes of varying lengths (100-500), obtained using the signature-based NN models with varying signature truncation orders (1-5), are averaged over 3 training runs. Here, $H\sim\text{Uniform}(0,\,1)$ for the fBm, $H\sim\text{Uniform}(0.5,\,1)$ for the fOU, and $H\sim\text{Beta}(1,\,9)$ for the rHeston. The best accuracies across different signature truncation orders are in bold.}
	\label{tab-sensitivity-truncation}
	\centering
	\smallskip
	\centering
	\renewcommand\arraystretch{1.5}
	\setlength{\tabcolsep}{3pt}
	\begin{tabular}{l c c c c c c c c}
		\toprule
		\multirow{2}{*}{Models}&\multicolumn{3}{c}{Test RMSE for input length $100$} &\multicolumn{3}{c}{Test RMSE for input length $500$} &\multirow{2}{*}{$\#$Params}&\multirow{2}{*}{Truncation order}\\
		\cline{2-4}\cline{5-7}
		& fBm & fOU & rHeston & fBm & fOU & rHeston & &\\
		\hline
		\multirow{3}{*}{DeepSigNet} &9.52e-2&1.34e-1&5.89e-2&9.32e-2&1.37e-1&5.92e-2&4461&1\\
		&1.72e-2&\textbf{3.34e-2}&5.81e-2&\textbf{8.41e-3}&\textbf{2.18e-2}&5.03e-2&9261&3\\
		&\textbf{1.34e-2}&3.41e-2&\textbf{5.78e-2}&4.53e-2&2.93e-2&\textbf{3.44e-2}&129261&5\\
		\hline
		\multirow{3}{*}{SigMA} &1.18e-1&1.34e-1&5.89e-2&1.26e-1&1.37e-1&5.92e-2&4726&1\\
		&1.51e-2&2.96e-2&5.44e-2&\textbf{8.89e-3}&1.59e-2&1.02e-2&87226&3\\
		&\textbf{1.06e-2}&\textbf{2.44e-2}&\textbf{3.48e-2}&8.24e-2&\textbf{1.26e-2}&\textbf{1.01e-2}&46024726&5\\
		\hline
		\multirow{3}{*}{SigSA} &2.85e-1&1.36e-1&1.15e-1&2.74e-1&1.39e-1&7.59e-2&15&1\\
		&1.32e-1&1.35e-1&5.89e-2&1.33e-1&1.39e-1&5.92e-2&603&3\\
		&\textbf{1.09e-1}&\textbf{1.35e-1}&\textbf{5.89e-2}&\textbf{1.08e-1}&\textbf{1.39e-1}&\textbf{5.92e-2}&11595&5\\
		\bottomrule
	\end{tabular}
	\end{minipage}}
\end{table}

For further explanations, we perform an interpretability analysis based on attention weights and MLP feature importance. 
Specifically, we introduce two metrics: a segment-wise index averages the softmax weights from query-key interactions to measure the contribution of different lifted segments, and a level-wise index uses the mean absolute weights of the MLP's first linear layer to measure the importance of different signature truncation levels. 	
The results, shown in Figure 3, suggest that both local and global segments contribute comparably to the prediction, while the most informative signature level appears around truncation order $N=3$. 
These observations are consistent with the approximate scale-invariant properties of the processes considered.

\begin{figure}[H]
	\centering
	\begin{subfigure}[b]{0.4\textwidth}
		\includegraphics[width=\textwidth]{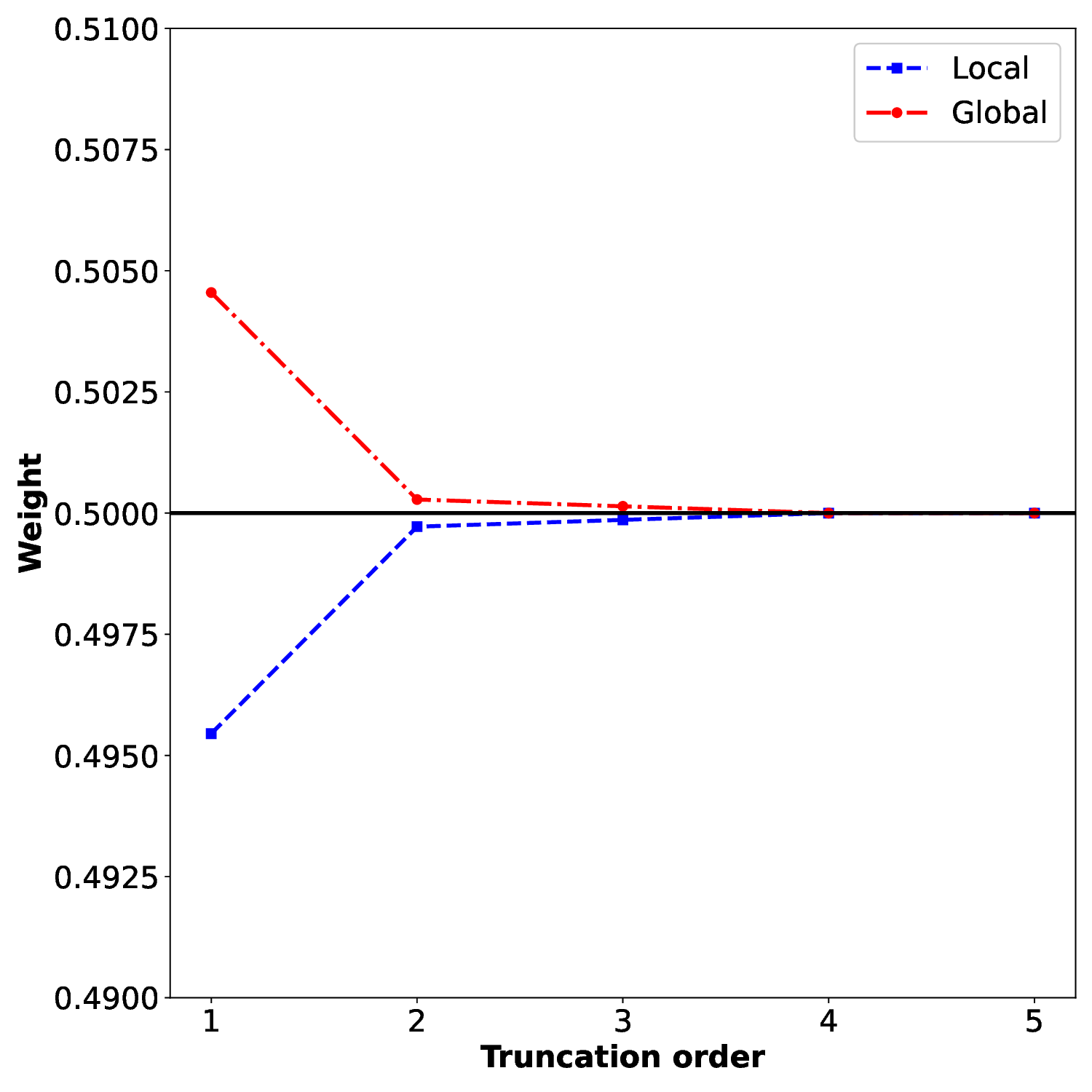}
		\caption{Segment-wise index}
	\end{subfigure}
	\hfill
	\begin{subfigure}[b]{0.4\textwidth}
		\centering
		\includegraphics[width=\textwidth]{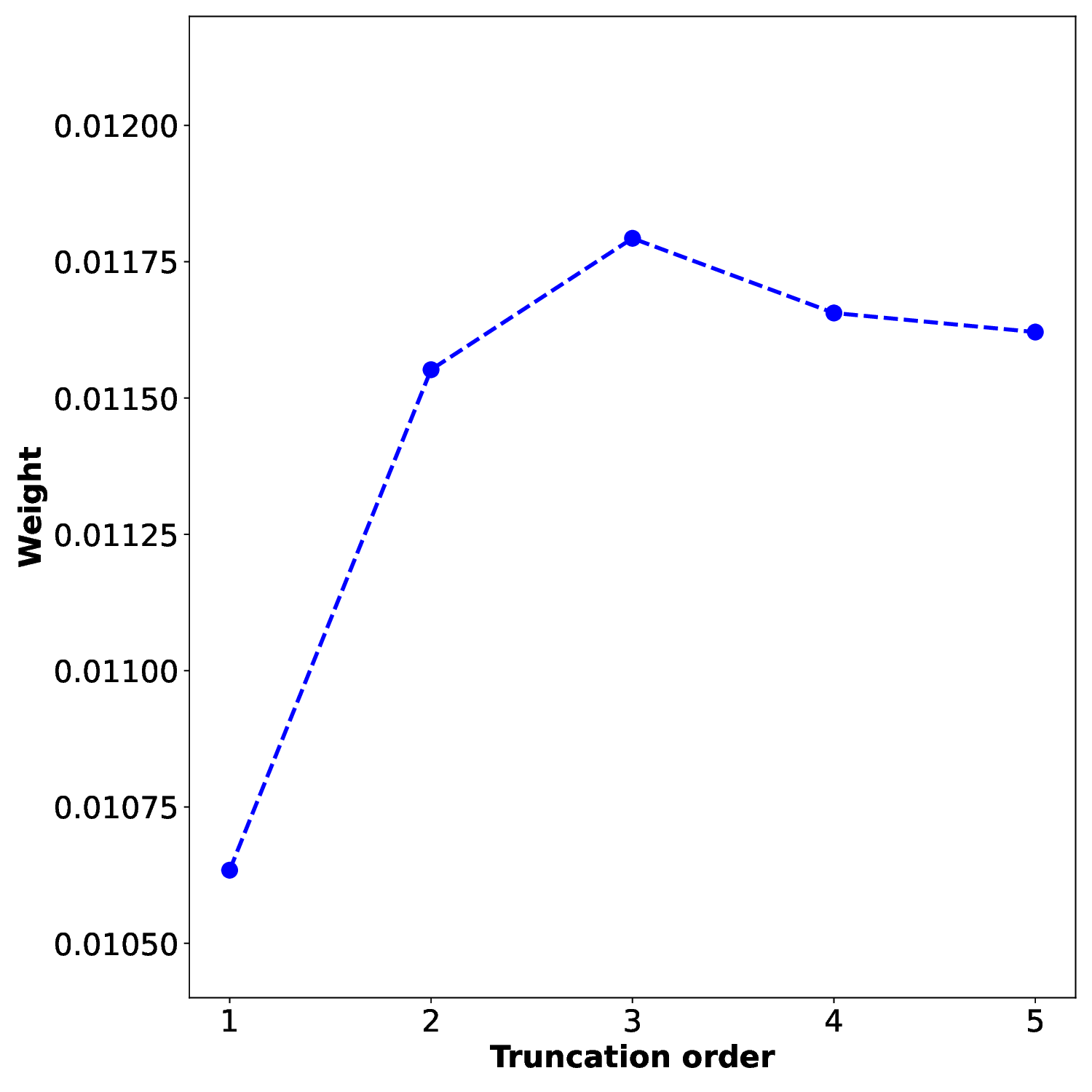}
		\caption{Level-wise index}
	\end{subfigure}
	\caption{Interpretability analysis for SigMA on rHeston paths with length 500. }
	\label{fig-sensitivity-truncation2}
\end{figure}

One potential concern is the computational cost of evaluating path signatures, since their mathematical definition involves iterated integrals (see Equation~\eqref{eq-signature}).
However, thanks to Chen's identity (Proposition~\ref{prop-chen}), the computation can be performed algebraically in practice, avoiding the need for explicit numerical integration. 
In particular, the signature of a piecewise linear path can be computed by first evaluating the signatures of linear segments between consecutive observations and then combining them recursively through tensor products. 
	
In our implementation, this procedure is carried out using the highly optimized \texttt{iisignature} package~\cite{reizenstein2020algorithm}, which provides efficient algorithms for computing truncated signatures and log-signatures. 
For a fixed truncation order, the computational cost of signature evaluation scales approximately linearly with the sequence length, making it practical for the trajectory lengths considered in our experiments.  

\subsubsection{Ablation: convolutional and MLP layers}
\label{subsubsec-sensitivity-structure}
In this section, we investigate the impact of the convolutional and MLP layers in the SigMA model.  It is noted in
\cite{tong_2023_SigFormer}  that the Transformer in their model can extract representations from signatures without using a convolutional layer. 
Therefore, we aim to determine whether including the convolutional and MLP layers in the SigMA model improves accuracy and stability during training. 
We compare SigMA to three different architectures, varying the inclusion of these layers (details in~\ref{apdix-architecture}). 

\begin{figure}
	\centering
	\includegraphics[width=0.6\textwidth]{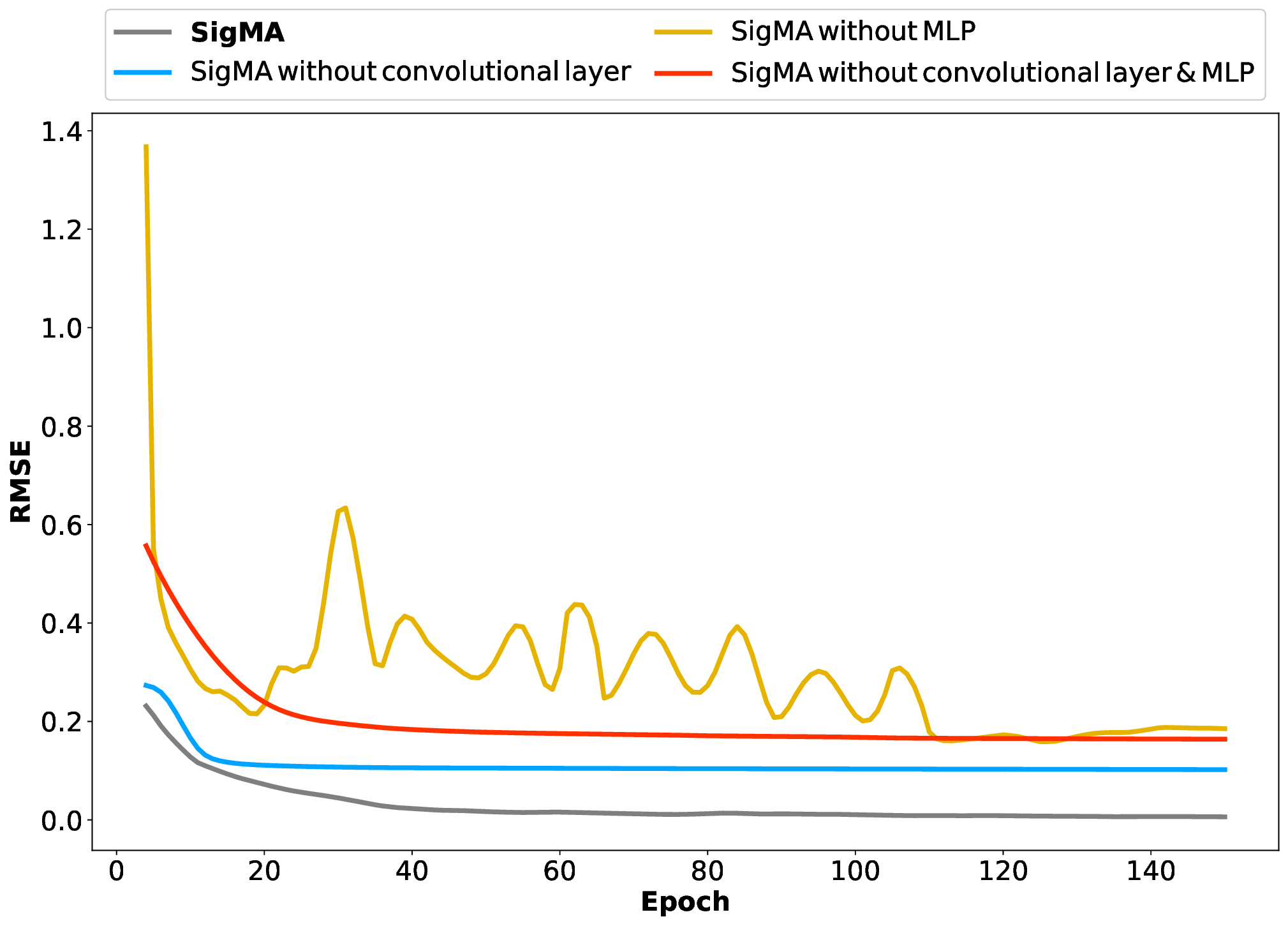}
	\caption{Loss curve from a typical training run for Hurst estimation of the fBm process with a sequence length of 500.}
	\label{fig-sensitivity-structure}
\end{figure}

The results are summarized in Table~\ref{tab-sensitivity-structure}, and a representative loss curve is shown in Figure~\ref{fig-sensitivity-structure}. 
These results demonstrate that the SigMA architecture outperforms the one used in \cite{tong_2023_SigFormer} (i.e., SigMA without convolutional layer \& MLP) in terms of test error accuracy. 
This superiority extends across various stochastic processes and different training data settings (i.e., length and sampling of $H$). 
Figure \ref{fig-sensitivity-structure} further suggests that SIGMA's architecture exhibits stable training dynamics. 
Notably, the inclusion of the MLP appears to contribute significantly to this robustness, making the SigMA model stable and accurate. Architectures without the MLP achieve good training and test accuracy in some cases but lack consistency and can be erratic. 
By contrast, adding a convolutional layer appears to improve training accuracy. 

We emphasize that the inclusion of the convolutional and MLP layers in SigMA is more than an architectural aggregation. 
They conceptually serve as an essential encoder and decoder, respectively. 
In particular, the convolutional layer acts as a learnable encoder that extracts local patterns and maps the raw trajectory into a latent representation. 
Additionally, as discussed in Remark~\ref{remark-convolutional-layer}, the convolutional layer also mitigates the "curse of dimensionality". 
The signature transform then provides a geometric representation of this latent path. 
Subsequently, the attention and MLP layers act as a decoder that maps these geometric features to parameter estimates. 
This "Encoder–Geometry–Decoder" pipeline fundamentally supports the stability and accuracy observed in our experiments. 

Finally, we performed a grid search over the convolutional kernel size and the hidden dimension of the MLP (see~\ref{apdix-sensitivity-structure-hyperparameter} for details) to verify that our results are robust to hyperparameter choices. 

\begin{table}
	\centering
	\centering
	\rotatebox{90}{
	\begin{minipage}{\textheight}
	\caption{Test RMSEs for Hurst estimation across stochastic processes of varying lengths (100-1500), obtained using the SigMA model and its variants, are averaged over 3 training runs. Here, $H\sim\text{Uniform}(0,\,1)$ for the fBm, $H\sim\text{Uniform}(0.5,\,1)$ for the fOU, and $H\sim\text{Beta}(1,\,9)$ for the rHeston. The best accuracies across SigMA and its variants are in bold.}
	\label{tab-sensitivity-structure}
	\centering
	\smallskip
	\renewcommand\arraystretch{1.1}
	\begin{threeparttable}
		\begin{tabular}{l c c c c c}
			\toprule
			\multirow{2}{*}{Models}&\multirow{2}{*}{Input lengths}&\multicolumn{3}{c}{Stochastic processes}&\multirow{2}{*}{$\#$Params}\\
			\cline{3-5}
			&&fBm&fOU&rHeston&\\
			\hline
			\multirow{4}{*}{SigMA}
			&100&\textbf{1.50e-2}&\textbf{2.81e-2}&5.20e-2&87226\\
			&500&\textbf{8.30e-3}&\textbf{1.68e-2}&\textbf{1.20e-2}&87226\\
			&1000&\textbf{1.03e-2}&\textbf{1.71e-2}&\textbf{7.23e-3}&87226\\
			&1500&\textbf{2.49e-2}&\textbf{2.20e-2}&\textbf{7.10e-3}&87226\\
			
			\hline
			\multirow{4}{*}{SigMA without convolutional layer}
			&100&1.06e-1&1.31e-1&5.88e-2&5647\\
			&500&1.05e-1&1.33e-1&5.91e-2&5647\\
			&1000&1.05e-1&1.34e-1&5.90e-2&5647\\
			&1500&1.02e-1&1.32e-1&5.99e-2&5647\\
			
			\hline
			\multirow{4}{*}{SigMA without MLP}
			&100&1.15e-1&7.08e-2&\textbf{3.80e-2}&73328\\
			&500&1.39e-1&7.73e-2&1.82e-2&73328\\
			&1000&1.80e-1&7.72e-2&1.83e-2&73328\\
			&1500&2.44e-1&7.93e-2&1.88e-2&73328\\
			
			\hline
			\multirow{4}{*}{SigMA  without convolutional layer \& MLP\tnote{1}}
			&100&1.68e-1&1.33e-1&5.85e-2&491\\
			&500&1.64e-1&1.36e-1&5.88e-2&491\\
			&1000&1.72e-1&1.35e-1&5.82e-2&491\\
			&1500&1.61e-1&1.33e-1&5.92e-2&491\\
			
			\bottomrule	
		\end{tabular}
		\begin{tablenotes}
			\footnotesize
			\item[1] SigMA without convolutional layer \& MLP actually has the same architecture as the SigFormer model used in \cite{tong_2023_SigFormer}.
		\end{tablenotes}
	\end{threeparttable}
	\end{minipage}}
\end{table}

\subsection{Hurst parameter estimation}
\label{subsec-experiments-single}
The analysis in Section \ref{subsec-experiments-sensitivity} provides a reference for the architecture of the SigMA model, which we will evaluate using the Hurst parameter estimation problem in this section. 
To this end, we train a variety of neural network models, as mentioned in the beginning of Section \ref{sec-experiments}, to perform the estimation using synthetic paths from Section \ref{subsubsec-sensitivity-implementation}. 
That is, we aim to learn the mapping $\mathbf{X}^{(n)}\mapsto H$. 
For comparison, we also introduce three classical estimators: the Detrended Fluctuation Analysis (DFA), the Wavelet estimator and the Whittle estimator. 
All the settings remain the same as described in Section \ref{subsubsec-sensitivity-implementation}. 

\begin{figure}
	\centering
	\begin{subfigure}[b]{0.32\textwidth}
		\includegraphics[width=\linewidth]{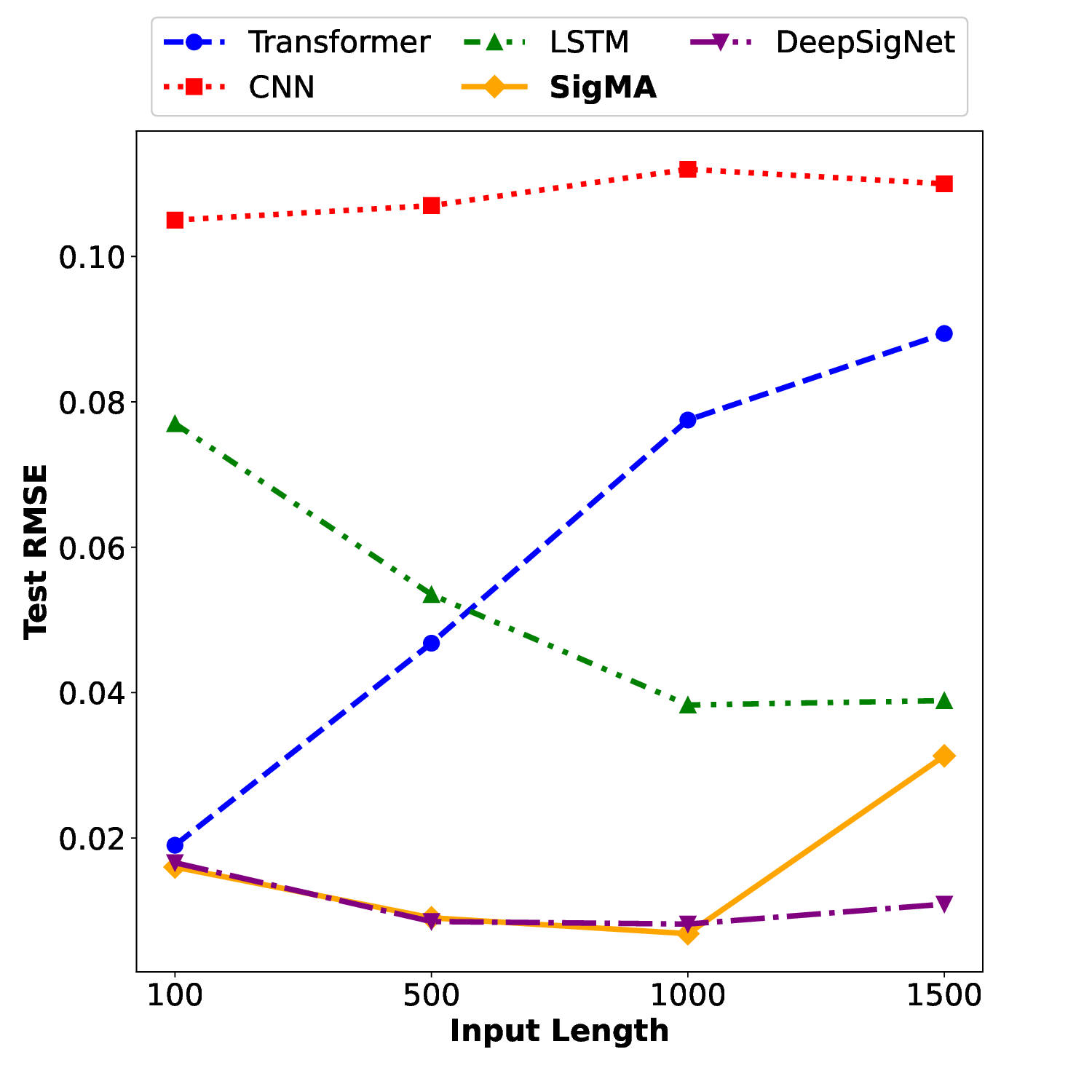}
		\caption{fBm}
		\label{fig-calibration-single-fBm}
	\end{subfigure}
	\hfill
	\begin{subfigure}[b]{0.32\textwidth}
		\includegraphics[width=\linewidth]{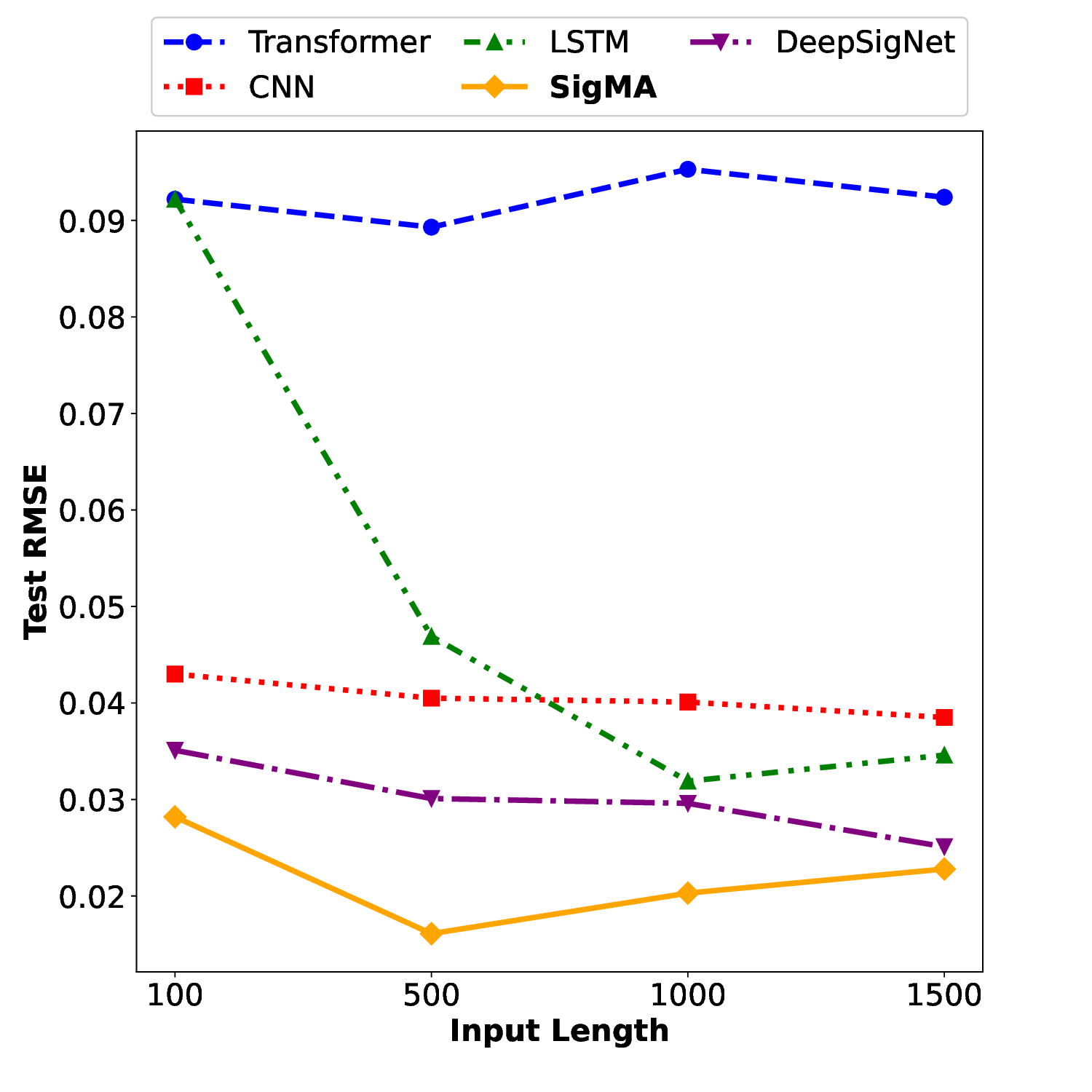}
		\caption{fOU}
		\label{fig-calibration-single-fOU}
	\end{subfigure}
	\hfill
	\begin{subfigure}[b]{0.32\textwidth}
		\includegraphics[width=\linewidth]{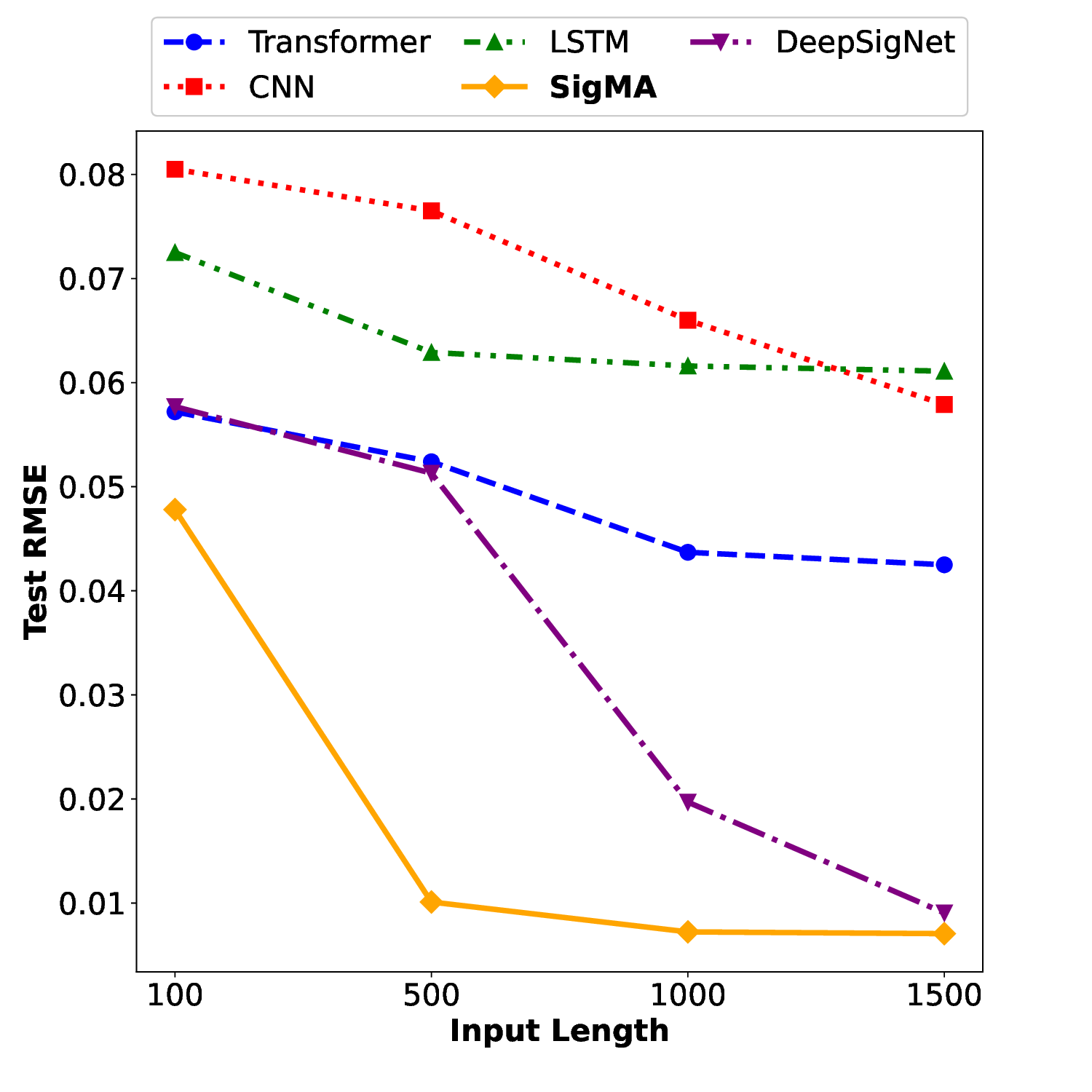}
		\caption{rHeston}
		\label{fig-calibration-single-rHeston}
	\end{subfigure}				
	\caption{Test RMSE trends for Hurst estimation across stochastic processes of varying lengths (100-1500), obtained using various NN models, are averaged over 3 training runs. Here, $H\sim\text{Uniform}(0,\,1)$ for the fBm, $H\sim\text{Uniform}(0.5,\,1)$ for the fOU, and $H\sim\text{Beta}(1,\,9)$ for the rHeston.}
	\label{fig-calibration-single}
\end{figure}

We begin this section by evaluating SigMA on the Hurst parameter estimation for one-dimensional fBm-driven processes. 
We present the results for both NN models and classical estimators in Table \ref{tab-calibration-single}, and plot only the NN results in Figure \ref{fig-calibration-single} to focus on the comparative performance of the neural architectures. 

\begin{table}
	\caption{Test RMSEs for Hurst estimation across stochastic processes of varying lengths ($100$-$1500$), obtained using various NN models and classical estimators, are averaged over 3 training runs. Here, $H\sim\text{Uniform}(0,\,1)$ for the fBm, $H\sim\text{Uniform}(0.5,\,1)$ for the fOU, and $H\sim\text{Beta}(1,\,9)$ for the rHeston. The best accuracies across different NN models are in bold.}
	\label{tab-calibration-single}
	\smallskip
	\centering
	\renewcommand\arraystretch{0.96}
	\setlength{\tabcolsep}{7pt}
	\begin{tabular}{l c c c c c}
		\toprule
		\multirow{2}{*}{Models}&\multirow{2}{*}{Input lengths}&\multicolumn{3}{c}{Stochastic processes}&\multirow{2}{*}{$\#$Params}\\
		\cline{3-5}
		&&fBm&fOU&rHeston&\\
		\hline
		\multirow{4}{*}{Transformer\tnote{1}}
		&100&1.90e-2&9.22e-2&5.72e-2&573906\\
		&500&4.68e-2&8.93e-2&5.24e-2&2557906\\
		&1000&7.75e-2&9.53e-2&4.37e-2&5037906\\
		&1500&8.94e-2&9.24e-2&4.25e-2&7517906\\
		\hline
		\multirow{4}{*}{CNN\tnote{2}}
		&100&1.05e-1&4.30e-2&8.05e-2&255073\\
		&500&1.07e-1&4.05e-2&7.65e-2&500833\\
		&1000&1.12e-1&4.01e-2&6.60e-2&812129\\
		&1500&1.10e-1&3.85e-2&5.79e-2&1107041\\
		\hline
		\multirow{4}{*}{LSTM\tnote{3}}
		&100&7.70e-2&9.22e-2&7.25e-2&1829634\\
		&500&5.35e-2&4.69e-2&6.29e-2&8383234\\
		&1000&3.83e-2&3.19e-2&6.16e-2&16575234\\
		&1500&3.89e-2&3.46e-2&6.11e-2&24767234\\
		\hline
		\multirow{4}{*}{SigMA}
		&100&\textbf{1.60e-2}&\textbf{2.82e-2}&\textbf{4.78e-2}&87226\\
		&500&9.02e-3&\textbf{1.61e-2}&\textbf{1.01e-2}&87226\\
		&1000&\textbf{6.84e-3}&\textbf{2.03e-2}&\textbf{7.23e-3}&87226\\
		&1500&3.13e-2&\textbf{2.28e-2}&\textbf{7.06e-3}&87226\\
		\hline
		\multirow{4}{*}{DeepSigNet\tnote{4}}
		&100&1.66e-2&3.51e-2&5.77e-2&9261\\
		&500&\textbf{8.51e-3}&3.01e-2&5.13e-2&9261\\
		&1000&8.16e-3&2.96e-2&1.97e-2&9261\\
		&1500&\textbf{1.09e-2}&2.51e-2&9.05e-3&9261\\
		\hline
		\multirow{4}{*}{DFA}
		&100&4.28e-01&2.61e-01&4.54e-01&-\\
		&500&3.79e-01&2.03e-01&2.72e-01&-\\
		&1000&3.88e-01&1.99e-01&2.37e-01&-\\
		&1500&3.86e-01&1.93e-01&2.23e-01&-\\
		\hline
		\multirow{4}{*}{Wavelet}
		&100&4.15e-01&2.59e-01&2.29e-01&-\\
		&500&3.80e-01&2.15e-01&1.91e-01&-\\
		&1000&3.61e-01&1.97e-01&2.90e-01&-\\
		&1500&3.51e-01&1.92e-01&2.81e-01&-\\
		\hline
		\multirow{4}{*}{Whittle}
		&100&4.40e-01&2.58e-01&2.70e-01&-\\
		&500&4.06e-01&2.22e-01&1.62e-01&-\\
		&1000&4.11e-01&2.15e-01&1.38e-01&-\\
		&1500&4.06e-01&2.07e-01&1.30e-01&-\\
		\bottomrule	
	\end{tabular}
\end{table}

Surprisingly, the classical estimators perform poorly on our synthetic datasets compared to the NN models. 
However, this can be explained mathematically: the DFA requires scale windows spanning multiple decades, and the Wavelet/Whittle estimators suffer from severe variance in their lowest-frequency bins when applied to short trajectories.
Moreover, the drift of the fOU and the stochastic volatility of the rHeston violate the strict global scaling and stationarity assumptions required by asymptotic statistical estimators, further limiting their performance.   

For the results of NN models, there are three conclusions to be drawn. 
Firstly, the SigMA model tends to achieve the best accuracy for the case of fOU and rHeston and it also achieve good accuracy, comparing with the competing models, for the case of fBm. 
Moreover, both the SigMA and DeepSigNet achieve high accuracy with significantly lower complexity than competing models, a performance improvement we attribute to the signature transform integrated into their architectures. 
In fact, the signature transform can extract core patterns from sequences, thereby filtering noise while preserving critical information for further processing. 
Secondly, both the SigMA and DeepSigNet models offer a significant advantage: their complexity, measured by the number of parameters, does not increase with input length. This excellent feature arises from the introduction of the signature transform, where the length of the path signature depends only on the number of features. 
Notably, the lifting operation before taking the signature transform would violate this advantage, therefore, as discussed in Section \ref{subsubsec-sensitivity-stride}, we choose the stride equal to half the input length to avoid this issue. 
Thirdly, for pure fBm the test error of SigMA slightly deteriorates when increasing the number of time steps from 1000 to 1500 (Figure \ref{fig-calibration-single-fBm}), a behaviour also observed for the Transformer baseline. 
We attribute this to the combination of the strong self-similarity of fBm and the fixed-dimensional signature/attention bottleneck in the architecture: as the path length grows, a larger number of highly correlated increments must be compressed into a signature of fixed order, so that additional observations become largely redundant and may even degrade generalization. 
Similar non-monotone finite-sample effects with respect to sample size have been reported for classical Hurst estimators in long-memory settings. 
By contrast, the fOU and rHeston models possess additional drift and volatility structure; for these processes longer paths carry genuinely more information, and SigMA’s performance continues to improve (at least does not deteriorate) as the input length increases (Figures \ref{fig-calibration-single-fOU}-\ref{fig-calibration-single-rHeston}).

\begin{figure}[H]
	\centering
	\includegraphics[width=0.8\textwidth]{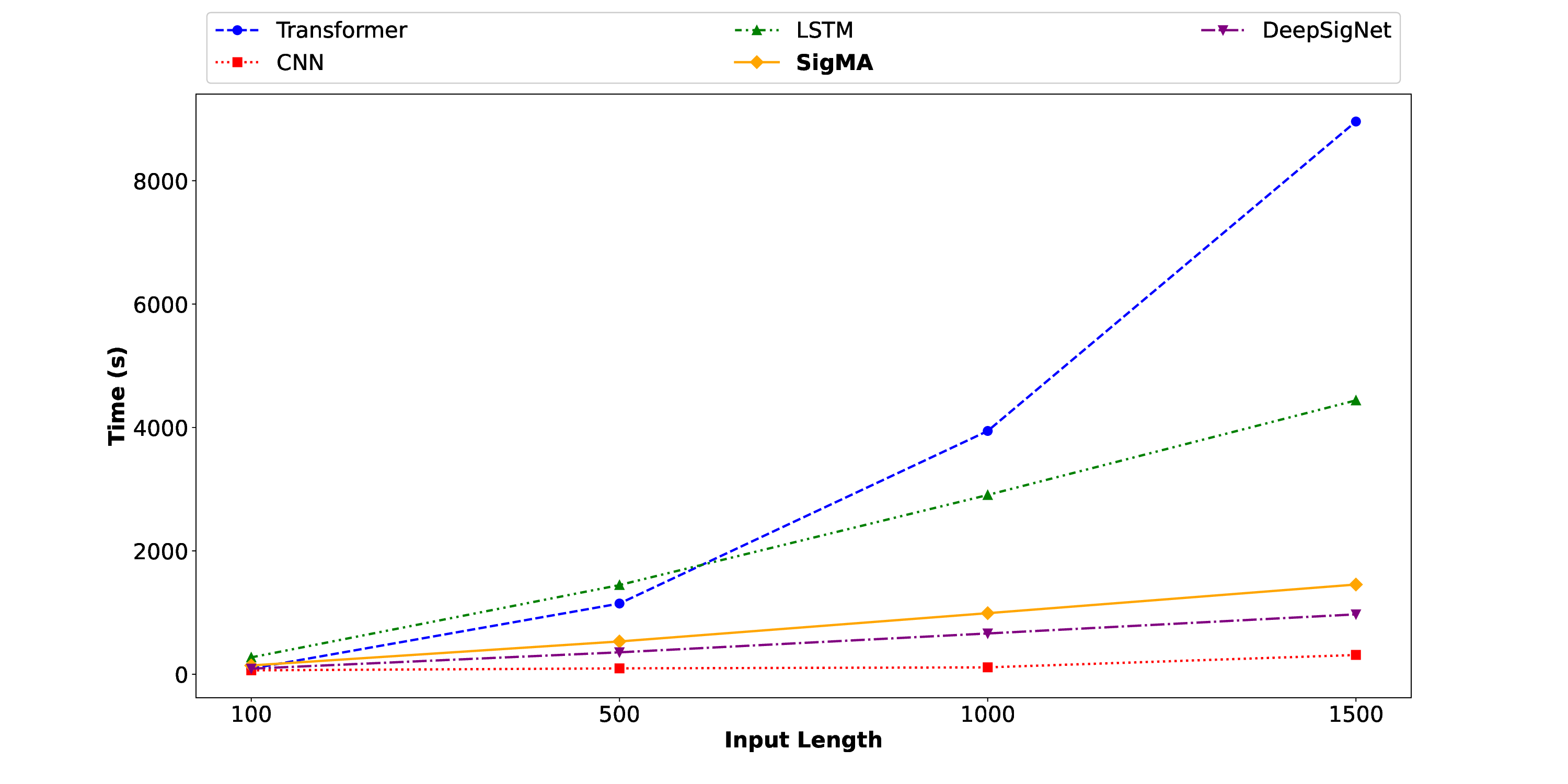}
	\caption{The training time (in seconds) of a typical run for Hurst estimation of the fOU process across all NN models. }
	\label{fig-calibration-single-time}
\end{figure}

As an illustration, we provide empirical measurements of computational cost for Hurst estimation of the fOU process. 
Figure~\ref{fig-calibration-single-time} reports the measured training time for all neural architectures as a function of the input trajectory length under identical experimental conditions. 
The results show that SigMA exhibits substantially lower training time than transformer-based and recurrent architectures, while remaining moderately more expensive than convolutional networks due to the additional preprocessing
step associated with signature computation. 

For fixed truncation order, this preprocessing overhead grows approximately linearly with the sequence length. 
Moreover, we observe that the computational cost of SigMA remains comparable to that of other sequence models and is
significantly more favorable than that of transformer architectures for the trajectory lengths considered in this study.

\begin{figure}[h]
	\centering
	\begin{subfigure}[b]{0.4\textwidth}
		\includegraphics[width=\textwidth]{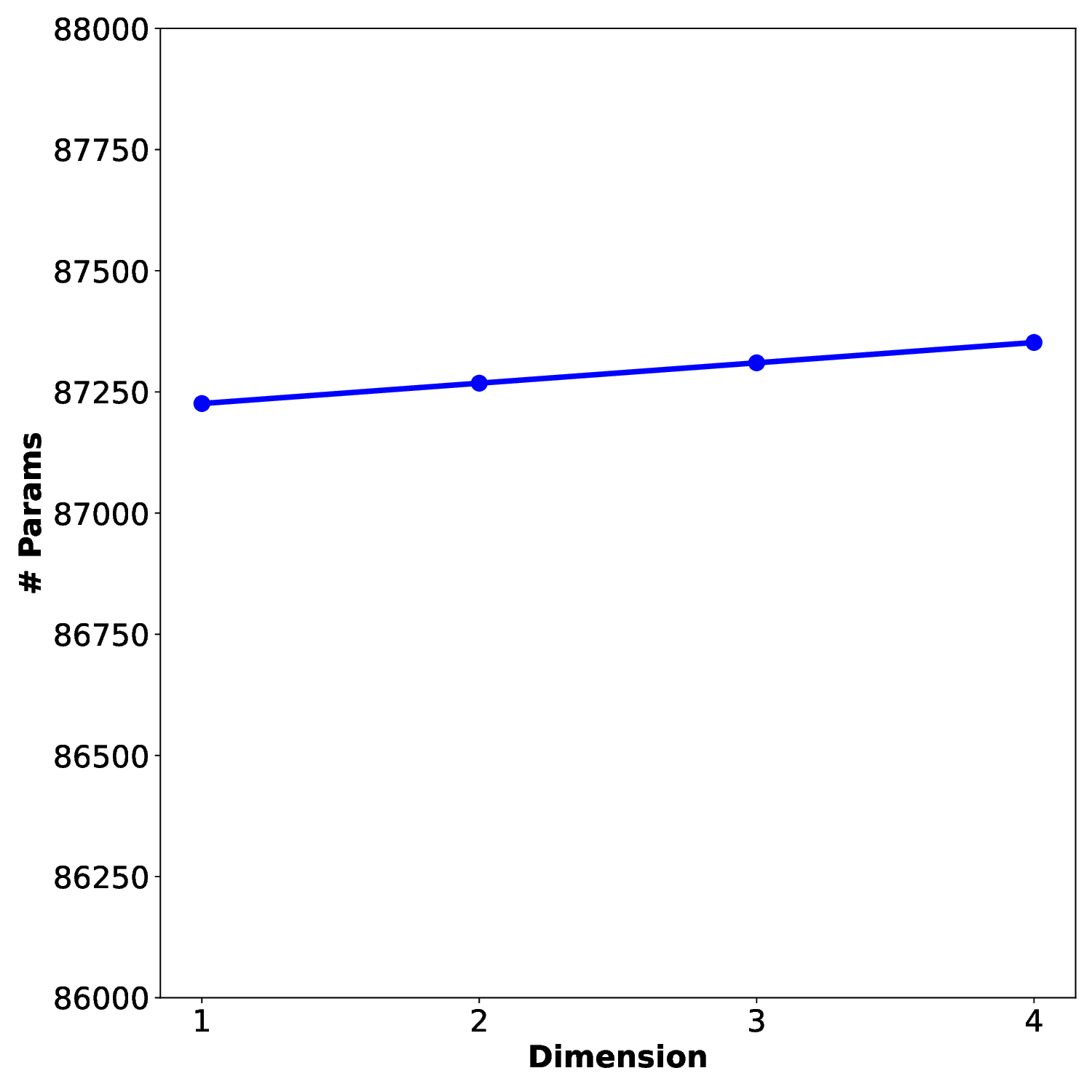}
		\caption{Parameter size vs. Dimensionality}
	\end{subfigure}
	\hfill
	\begin{subfigure}[b]{0.4\textwidth}
		\centering
		\includegraphics[width=\textwidth]{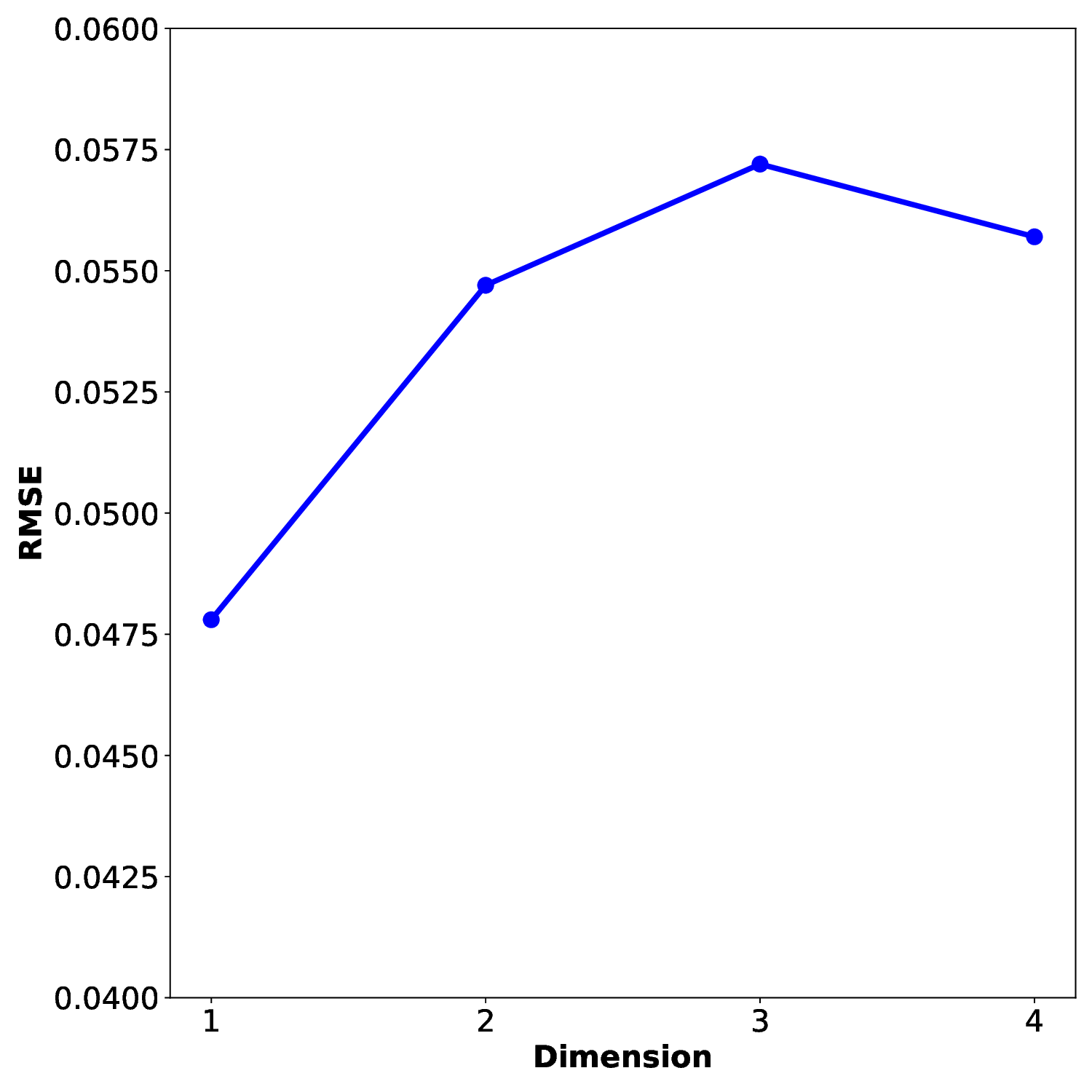}
		\caption{RMSE vs. Dimensionality}
	\end{subfigure}
	\caption{Performance of SigMA on multivariate fOU processes across different dimensions. }
	\label{fig-calibration-multivariate}
\end{figure}

Next, we test SigMA on multivariate fOU processes driven by multiple fBms \cite{dugo2025multivariate} to explore its performance and scalability in higher dimensions. 
Here, we use SigMA to simultaneously estimate the Hurst parameters of multiple fBms driving the multivariate fOU processes. 
The results in Figure~\ref{fig-calibration-multivariate} verify that the input dimension has little impact on SigMA's scalability, thanks to the convolutional preprocessing layer discussed in Remark~\ref{remark-convolutional-layer}. 
While the accuracy decreases slightly in the multivariate setting, it stays well within an acceptable limit. 

\subsection{Joint multi-parameter estimation}
\label{subsec-experiments-multiple}
In the previous example, we observed that the SigMA model tends to outperform competing models in many complex single-parameter estimation problems. Therefore, it is natural to consider extending the SigMA model to multiple-parameter estimation problems. 

\subsubsection{Implementation details}
\label{subsubsec-multiple-implementation}
To generate the datasets for multiple-parameter estimation problems, we first sample different values of $\alpha,\,\mu,\,\sigma,\,\kappa_1,\,\kappa_2$ and $\theta$ from the distribution shown in Table \ref{tab-multiple-distribution}. 
We then simulate fOU and rHeston paths using these parameters, along with $H\sim\text{Uniform}(0.5,\,1)$ and $H\sim\text{Beta}(1,\,9)$, respectively, to generate two corresponding training/test datasets of the same size as shown in Section \ref{subsubsec-sensitivity-implementation}, and all these paths are simulated with $500$ steps. 
The first dataset consists of fOU paths with varying values of $\alpha,\,\mu,\,\sigma$ and $H$, while the second dataset contains rHeston paths with varying $\kappa_1,\,\kappa_2,\,\theta$ and $H$. 
All other settings remain the same as in Section \ref{subsubsec-sensitivity-implementation}.
\begin{table}[H]
	\caption{Distributions for sampling parameters.}
	\label{tab-multiple-distribution}
	\smallskip
	\centering
	\renewcommand\arraystretch{1.2}
	\setlength{\tabcolsep}{25pt}
	\begin{tabular}{l c}
		\toprule
		Parameters&Distributions\\
		\hline
		$\alpha$&$\text{Uniform}(0,\,5)$\\
		$\mu$&$\text{Uniform}(0,\,0.5)$\\
		$\sigma$&$\text{Uniform}(0,\,3)$\\
		$\kappa_1$&$\text{Uniform}(0,\,5)$\\
		$\kappa_2$&$\text{Uniform}(0,\,3)$\\
		$\theta$&$\text{Uniform}(0,\,0.5)$\\
		\bottomrule
	\end{tabular}
\end{table}

\subsubsection{Numerical results}
\label{subsubsec-multiple-calibration}
We perform two different multiple-parameter estimation experiments: the first involves simultaneous estimation of four parameters $H,\,\alpha,\,\mu$ and $\theta$ for the fOU process, and the second involves simultaneous estimation of $H,\,\kappa_1,\,\kappa_2$ and $\theta$ for the rHeston process. 
These two experiments represent joint estimation under conditions of long-range dependence and extreme roughness, respectively. 
For the sake of exposition, we first report the average RMSE\footnote{Let $\{(x_i,y_i)_{i=1,\ldots,n}:x_i,y_i\in\mathbb{R}^d\}$ be a dataset of $n$ samples, each with $d$ features. The average RMSE is then defined as $\frac{1}{d}\sqrt{\frac{1}{n}\sum\limits_{i=1}^n\|x_i-y_i\|_2^2}$, where $\|\cdot\|_2$ denotes the Euclidean norm.} for all parameters combined. 
Furthermore, to compare the robustness of these models, we also compute the ranked average root squared error (RSE)\footnote{Similarly, the average RSE of each sample is defined as $\frac{1}{d}\|x_i-y_i\|_2$ for $i=1,\ldots,n$.} on the test dataset samples, presenting the maximum, $75\%$ quantiles and $25\%$ quantiles of the ranked average RSE for all parameters. 
All the results are shown in Table~\ref{tab-multiple}. 

\begin{table}
	\centering
	\rotatebox{90}{
		\begin{minipage}{\textheight}
			\caption{Test results for multiple parameters estimation across stochastic processes with length 500, obtained using various NN models, are based on $10$ training runs. Here, beyond the average RMSEs for all samples, we also rank the average RSEs for every samples in test data set and present the maximum, $75\%$ quantiles and $25\%$ quantiles, respectively. The best accuracies across different NN models are in bold and models are sorted by ascending average RMSEs.}
			\label{tab-multiple}
			\smallskip
			\centering
			\renewcommand\arraystretch{1.5}
			\setlength{\tabcolsep}{12pt}
			\begin{tabular}{l c c c c c c}
				\toprule
				\multirow{2}{*}{Models}&\multirow{2}{*}{Stochastic processes}&\multirow{2}{*}{Average RMSEs}&\multicolumn{3}{c}{Average RSEs}&\multirow{2}{*}{$\#$Params}\\
				&&&max&q75\%&q25\%&\\
				\hline
				SigMA&\multirow{5}{*}{fOU}&\textbf{0.416}&\textbf{1.360}&\textbf{0.399}&\textbf{0.207}&87341\\
				Transformer&&0.436&3.830&0.408&0.208&2558005\\
				CNN&&0.489&1.388&0.506&0.275&501220\\
				LSTM&&0.594&2.369&0.598&0.278&8383299\\
				DeepSigNet&&0.855&21.757&0.545&0.298&9360\\
				\hline
				SigMA&\multirow{5}{*}{rHeston}&\textbf{0.391}&1.286&\textbf{0.402}&\textbf{0.161}&87341\\
				DeepSigNet&&0.398&1.332&0.415&0.164&9360\\
				Transformer&&0.406&1.258&0.413&0.168&2558005\\
				CNN&&0.499&\textbf{1.061}&0.525&0.247&501220\\
				LSTM&&0.612&1.691&0.634&0.312&8383299\\
				\bottomrule
			\end{tabular}
	\end{minipage}}
\end{table}

In both multiple-parameter estimation problems, the SigMA model achieves the best average RMSE with relatively fewer model parameters. 
Notably, the DeepSigNet exhibits an abnormally large maximum average RSE in estimating the fOU process, indicating the presence of outliers. 
This suggests that the model may have difficulty handling certain extreme cases in the data, leading to greater result variability. 
By contrast, the SigMA model exhibits superior consistency across all percentiles, demonstrating its enhanced robustness and reduced risk of overfitting in diverse scenarios.

\begin{figure}[H]
	\centering
	\begin{subfigure}[b]{0.49\textwidth}
		\includegraphics[width=\textwidth]{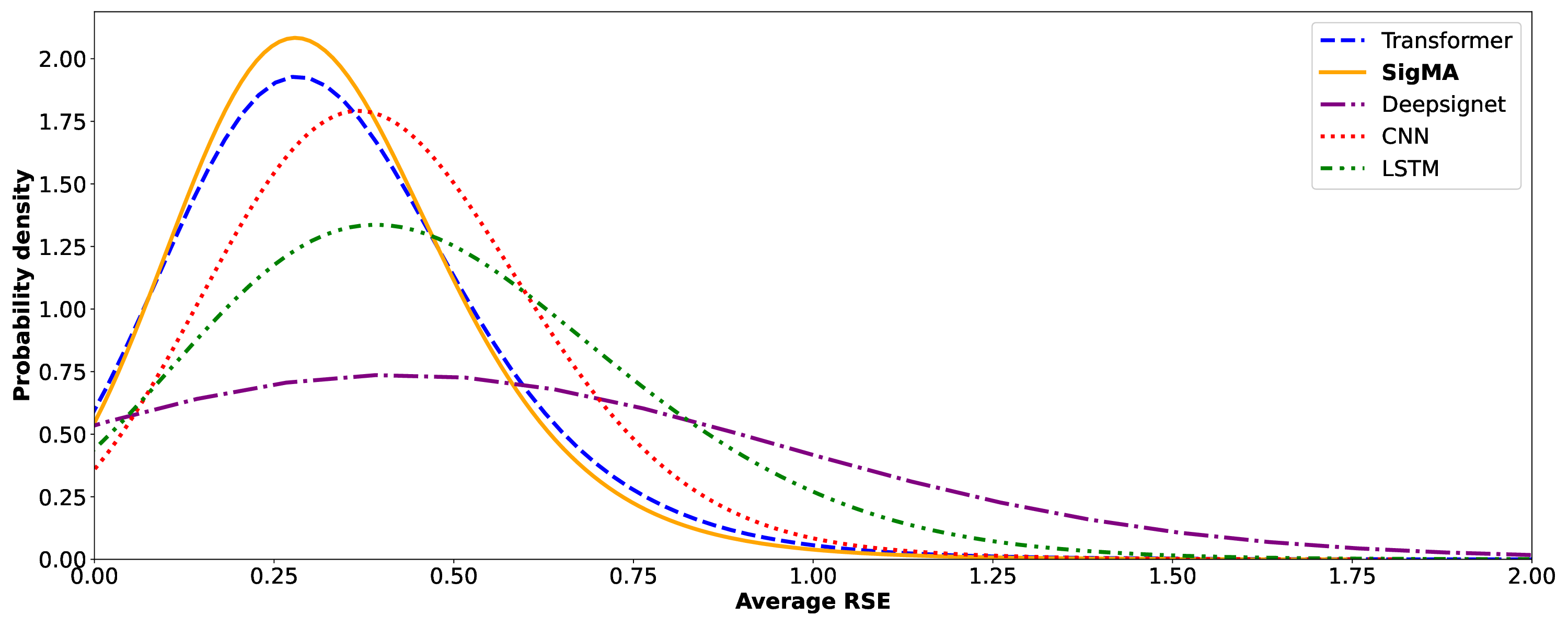}
		\caption{fOU}
	\end{subfigure}
	\hfill
	\begin{subfigure}[b]{0.49\textwidth}
		\centering
		\includegraphics[width=\textwidth]{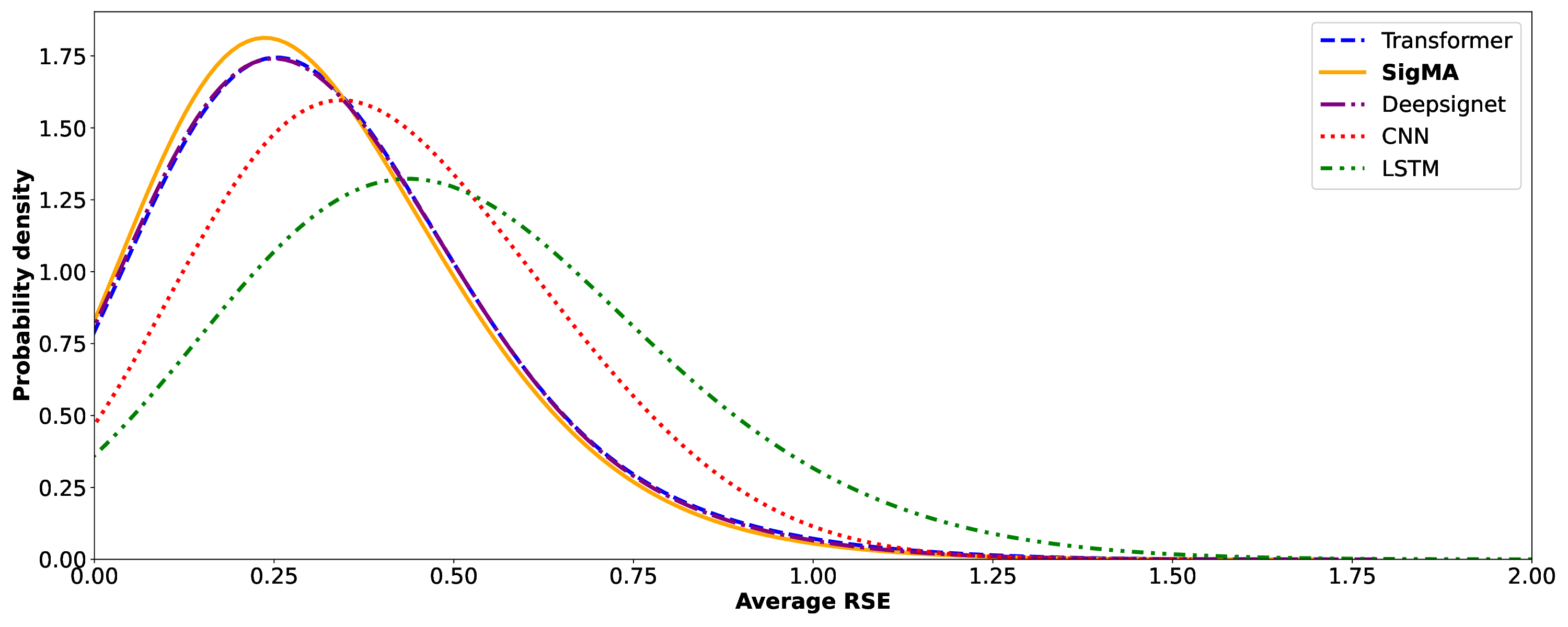}
		\caption{rHeston}
	\end{subfigure}
	\caption{Probability density distribution of average RSEs for multiple parameters estimation across stochastic processes with length 500, obtained using various NN models, are based on $10$ training runs. }
	\label{fig-multiple}
\end{figure}

Moreover, Figure \ref{fig-multiple} presents the distribution of the average RSEs' probability density, providing an overview of the errors. 
The higher peak of the SigMA model’s probability density function (PDF) indicates a more concentrated error distribution, demonstrating greater robustness than competing models.
Additionally, the leftward shift of the PDF peak suggests that SigMA attains a smaller median error relative to the other models. 
However, the DeepSigNet, which also employs a signature transform, exhibits considerable instability and low accuracy in estimating the fOU. 
This demonstrates that the additional structure of the SigMA effectively enhances the stability and accuracy of joint estimations. 
Ultimately, the SigMA model achieves these superior results while using significantly fewer parameters than the Transformer, LSTM and CNN model, making it an appealing choice for problems where computational efficiency and model simplicity are crucial. 

For further insight into the joint estimation results, Table~\ref{tab-multiple2} provides the RMSE and variance for each individual parameter. 
These results reveal that parameters governing short-term fluctuations (such as volatility-related parameters and the Hurst exponent) tend to be estimated more accurately than drift or mean-reversion parameters. 
A possible explanation is that volatility and roughness properties primarily influence local increments of the trajectory, which are effectively captured by low-order signature features. 
In contrast, drift and mean-reversion parameters influence the longer-term evolution of the process and therefore require longer observation horizons for accurate estimation~\cite{merton1980estimating, kleptsyna2002statistical, hu2010parameter}.

\subsection{Empirical applications on real-world datasets}
\label{subsec-experiments-empirical}
\subsubsection{Calibration to equity-index realized volatility}
\label{subsubsec-experiments-empirical-stock}
All the previous examples are based on simulated paths, in this example, we follow \cite{stone2020calibrating} to calibrate the H$\ddot{\rm o}$lder  (Hurst) exponent of historic realized volatility data from the Oxford–Man Institute of Quantitative Finance. 
Since the log-volatility behaves essentially as a fBm with the Hurst exponent $H$ of order $0.1$ (see \cite{gatheral2018volatility}), we generate the training and evaluation datasets using the rHeston process with $H\sim\text{Beta}(1,\,9)$, and set the sequence length to $100$, as illustrated in Section \ref{subsubsec-sensitivity-implementation}. 
We take a sample of $10$ different indices; for each index we then used a time series of $200$ sequential data points to create $11$ vectors of length $100$ (entries $1$ to $100$, $11$ to $110$ and so on) to predict the H$\ddot{\rm o}$lder exponent for each index. 
We compute the RMSE between the models' predictions and the least squares prediction, and the standard deviation of the difference between these predictions, see Table~\ref{tab-empirical-stock}. The SigMA model achieves the lowest RMSE and a favorable standard deviation among all models, while maintaining a reasonable level of model complexity. By contrast, although other models also show reasonable accuracy, they either require much more parameters or exhibit higher errors and standard deviations, indicating a trade-off between complexity and performance.
Therefore, we can state that the SigMA model is precise enough to be used in practice, offering an efficient and reliable solution for real-world applications. 

\begin{table}[H]
	\caption{H$\ddot{\rm o}$lder exponent estimation results for historic realized volatility data, obtained using various NN models. The best results across different NN models are in bold and models are sorted by ascending RMSEs. }
	\label{tab-empirical-stock}
	\smallskip
	\centering
	\renewcommand\arraystretch{1.2}
	\setlength{\tabcolsep}{15pt}
	\begin{tabular}{l c c c}
		\toprule
		\multirow{2}{*}{Models}&\multicolumn{2}{c}{Errors}&\multirow{2}{*}{$\#$Params}\\
		\cline{2-3}
		&RMSE&Std.&\\
		\hline
		SigMA&\textbf{1.75e-2}&3.07e-4&87226\\
		\hline
		Transformer&2.05e-2&\textbf{2.59e-4}&573906\\
		\hline
		DeepSigNet&3.48e-2&6.17e-4&9261\\
		\hline
		LSTM&3.66e-2&9.77e-4&1829634\\
		\hline
		CNN&6.97e-2&2.54e-3&255073\\
		\bottomrule
	\end{tabular}
\end{table}
\subsubsection{Application to Li-ion battery degradation}
\label{subsubsec-experiments-empirical-battery}
\begin{figure}[h]
	\centering
	\includegraphics[width=0.6\textwidth]{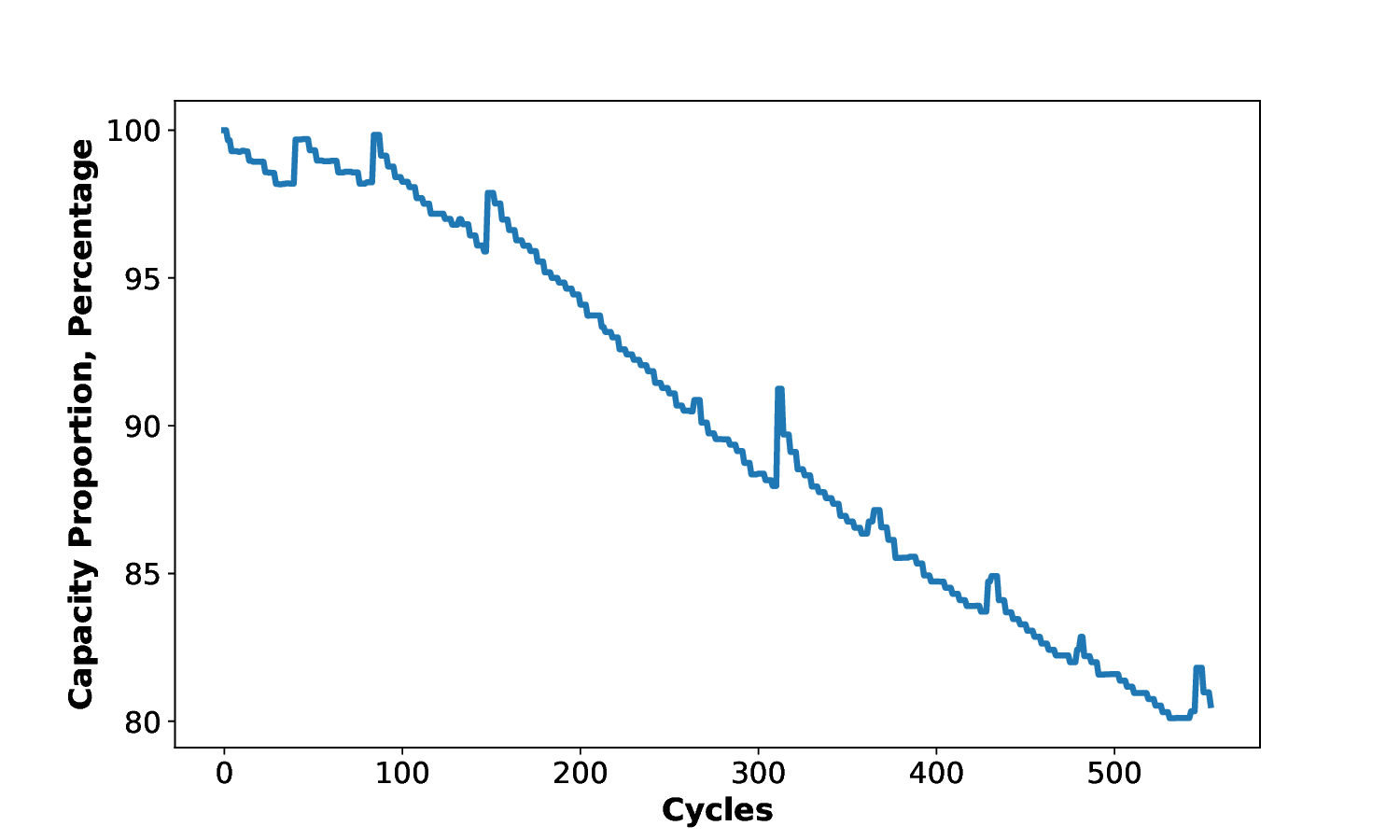}
	\caption{Capacity losses caused by structural degradation of Li-ion batteries. }
	\label{fig-battery-degradation}
\end{figure}

In this example, we consider the degradation of Li-ion batteries over charge-discharge cycles using capacity loss data from the NASA Prognostics Center of Excellence (PCoE) repository.\footnote{\url{https://www.nasa.gov/  intelligent-systems-division/discovery-and-systems-health/pcoe/pcoe-data-set-repository/}} We present the obtained series data in Figure~\ref{fig-battery-degradation}. 

In order to quantify the long-range dependence in the degradation dynamics, we use the Higuchi method 
and the rescaled range (R/S) analysis for statistical benchmarks \cite{csanady2024parameter}. 
Similar to Section~\ref{subsubsec-experiments-empirical-stock}, we use the series data of the battery capacity proportion to create $46$ vectors of length $100$-cycles (entries $0$-cycle to $99$-cycle, $10$-cycle to $109$-cycle and so on) to predict the Hurst parameter. 	

\begin{figure}[H]
	\centering
	\includegraphics[width=0.9\textwidth]{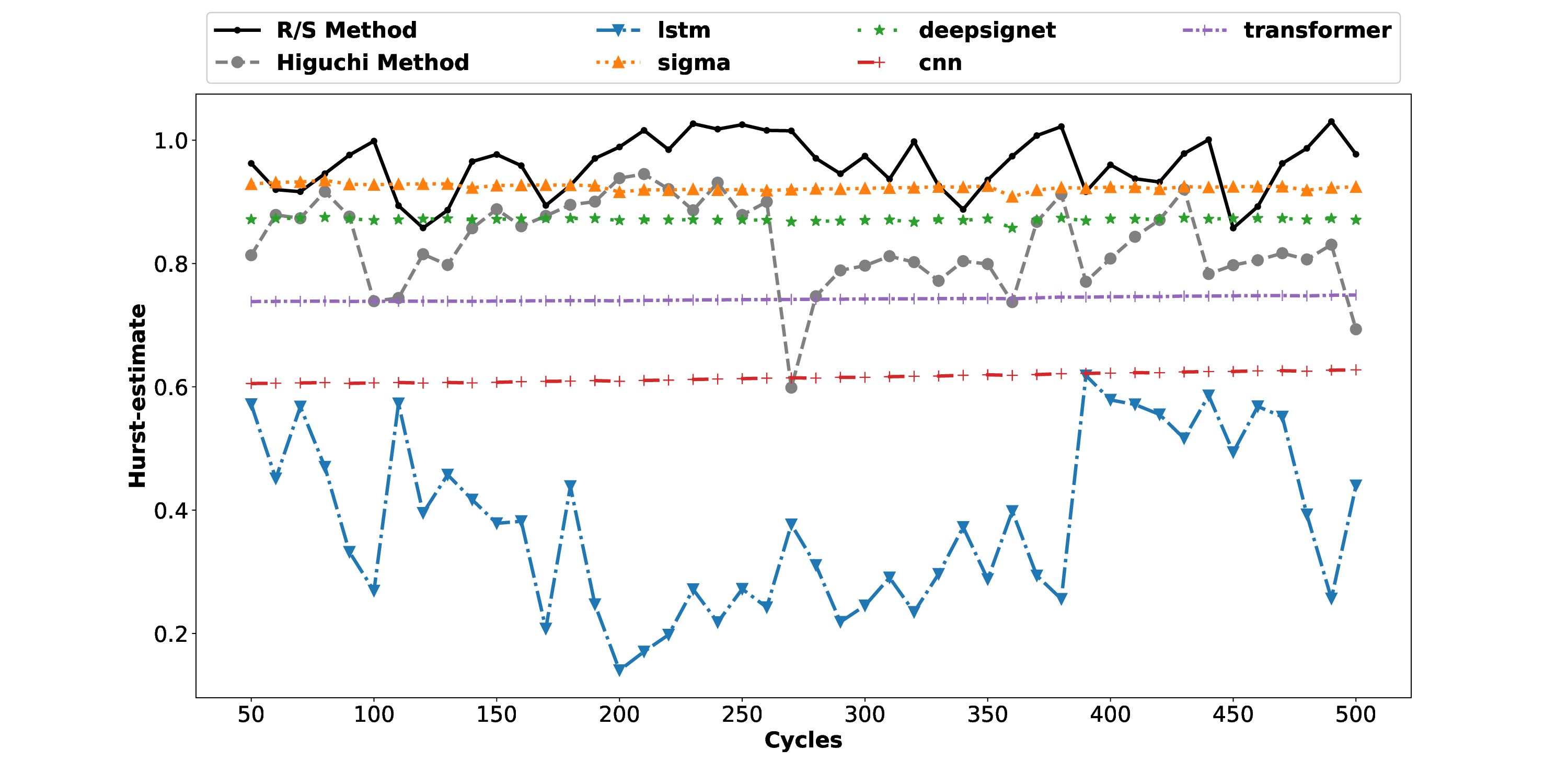}
	\caption{Hurst exponent estimates for battery capacity degradation series (50 to 500 cycles), obtained from NN models and statistical benchmarks. Results are based on 100-cycle sliding windows with 90-cycle overlap.}
	\label{fig-battery-estimation}
\end{figure}

The degradation series in Figure \ref{fig-battery-degradation} exhibits a global downward trend, local step-changes, and several regime shifts. 
Such behaviour violates the stationarity and self-similarity assumptions that underlie classical Hurst-exponent analysis. 
Consequently, the H-values reported in Table \ref{tab-empirical-battery} and Figure \ref{fig-battery-estimation} should be interpreted as local indicators of persistent path memory rather than as true Hurst exponents in the sense of fractional Brownian motion.  

The Higuchi method yields a mean value of $0.8330$, while the R/S analysis resulted in $0.9598$, indicating persistent long-memory behavior in the degradation process. 

Assuming the series data of the battery capacity proportion follows fBm, based on the obtained data series, we employ the neural networks trained by the fBm paths with $H\sim\text{Uniform}(0,1)$ to estimate its Hurst parameter. 
The results, together with statistical benchmarks, are shown in Table~\ref{tab-empirical-battery} and Figure~\ref{fig-battery-estimation}. The closest result to the R/S benchmark (0.9598) is from the SigMA (0.9239), while the closest result to Higuchi benchmark (0.8330) is from the DeepSigNet (0.8712). The results from LSTM and CNN exhibit significant deviations from both statistical benchmarks.

Across sliding windows of 100 cycles (with 90-cycle overlap), SigMA and DeepSigNet provide stable and relatively smooth estimates, consistently ranging between the Higuchi and R/S values. This behaviour is expected as:
\begin{itemize}
	\item the R/S method is well-known to overestimate H in the presence of linear or slowly varying trends, which explains the values close to 1 observed here;
	\item the Higuchi estimator, while less trend-sensitive, exhibits substantial fluctuations on short windows and irregular signals;
	\item by contrast, the signature-based ML models primarily exploit the structure of local increments and are therefore less affected by non-stationarity and short-window artifacts than classical methods. 
\end{itemize}
Among neural architectures, LSTM and CNN produce highly variable estimates with large cycle-to-cycle jumps, whereas SigMA and DeepSigNet yield the most stable behaviour. 
The Transformer baseline tends to underestimate H and remains almost flat, likely reflecting its tendency to average over the strong underlying trend. 
It is important to note that the choice of window length and overlap materially affects all estimators: using 100-cycle windows provides limited scaling information, and the 90\% overlap induces high autocorrelation between adjacent estimates. 
Longer or multiscale windows, or detrended fluctuation methods, could improve interpretability and robustness but fall outside the present scope. 
Nevertheless, the results demonstrate that signature-based learning, in particular SIGMA,  captures  persistent path memory properties of degradation trajectories more faithfully than conventional deep learning models. 

The degradation series exhibits strong non-stationarity, including long-term trends and local regime shifts. 
Classical estimators such as R/S analysis and Higuchi’s method rely on global scaling assumptions and are therefore sensitive to these effects. 
For example, the R/S estimator is known to overestimate the Hurst exponent in the presence of deterministic trends.
Signature-based models operate differently because they rely on geometric representations of path increments. 
The signature transform captures the ordered structure of increments rather than absolute values of the signal. 
In addition, the convolutional preprocessing layer extracts local patterns from the raw trajectory before the signature transform is applied. 
This combination makes the model less sensitive to isolated spikes or local irregularities in the observations.

\begin{table}[H]
	\caption{Hurst estimation results for the battery capacity proportion series from NN models and statistical benchmarks (in bold), sorted by ascending H-value.}
	\label{tab-empirical-battery}
	\smallskip
	\centering
	\renewcommand\arraystretch{1.1}
	\setlength{\tabcolsep}{6pt}
	\begin{tabular}{l c c c}
		\toprule
		Models/Methods & H-estimates & 95\% confidence intervals&$\#$Params\\
		\hline
		LSTM & 0.3779 & (0.3373, 0.4183) & 1829634\\
		\hline
		CNN & 0.6151 & (0.6129, 0.6172) & 255073\\
		\hline
		Transformer & 0.7425 & (0.7414, 0.7434) & 573906\\
		\hline
		\textbf{Higuchi} & \textbf{0.8330} & \textbf{(0.8120, 0.8539)} & -\\
		\hline
		DeepSigNet & 0.8712 & (0.8704, 0.8720) & 9261\\
		\hline
		SigMA & 0.9239 & (0.9224, 0.9252) & 87226\\
		\hline
		\textbf{R/S} & \textbf{0.9598} & \textbf{(0.9460, 0.9735)} & -\\			
		\bottomrule
	\end{tabular}
\end{table}

\subsection{Summary of architectural comparisons}
\label{subsec-experiments-summary}
Based on the results presented in Sections \ref{subsec-experiments-sensitivity}-\ref{subsec-experiments-empirical}, we summarize the differences between SigMA and comparable models in terms of architectural designs, applicable scenarios, and performance.

SigFormer applies attention mechanisms directly to signature features and uses a linear output layer. 
In contrast, SigMA introduces a convolutional preprocessing layer before the signature transform and an MLP after the attention layer. 
Empirically, we observe that replacing the final linear layer with an MLP improves both training stability and predictive accuracy (see Figure~\ref{fig-sensitivity-structure}), suggesting that the mapping from geometric signature features to model parameters is highly nonlinear. 

DeepSigNet directly feeds signature features into an MLP, whereas SigMA introduces a multi-head self-attention layer between the signature transform and the MLP. 
In our architecture, different attention heads process signature features associated with different truncation levels, allowing the model to weight geometric information arising from different interaction orders. 
In the joint multi-parameter estimation (Section~\ref{subsec-experiments-multiple}), DeepSigNet tends to over-compress the geometric information, leading to more outliers and higher quantiles (see Table~\ref{tab-multiple}). 
In contrast, by weighting the importance of different signature levels, SigMA can more effectively capture different geometric features of paths, leading to lower estimation errors and higher stability.

Finally, classical models such as CNN, LSTM, and Transformer operate directly on raw paths instead of path signatures. 
For long sequences, these models need significantly more parameters to capture temporal dependencies, making them expensive to train. 
As shown in Table~\ref{tab-calibration-single}, for an input length of $1500$ steps, the LSTM and Transformer models require over $24$M and $7.5$M parameters, respectively. 
In contrast, SigMA keeps a small, constant parameter count ($87,226$) regardless of the sequence length, achieving a better balance between model complexity and accuracy. 

\section{Conclusion}
\label{sec-conclusion}
In this work, we proposed SigMA (Signature Multi-head Attention), a neural architecture for parameter estimation in stochastic differential equations driven by fractional Brownian motion. Motivated by the non-Markovian and non-semimartingale nature of fBm-driven models, which limits the applicability of classical inference techniques, SigMA combines path signature representations with multi-head self-attention, complemented by convolutional preprocessing and multilayer perceptron components. This design explicitly targets the extraction of informative, path-level features while maintaining a compact and scalable model structure.

Through extensive numerical experiments, we demonstrated that SigMA achieves consistently strong performance across a range of estimation tasks, including Hurst parameter estimation and joint multi-parameter inference for fBm, fractional Ornstein–Uhlenbeck, and rough Heston models. In particular, the integration of signature transforms enables SigMA to decouple model complexity from input path length, while the attention mechanism effectively aggregates information across signature levels. Sensitivity and ablation studies further confirmed that the convolutional and MLP components play an important role in improving both accuracy and training stability. Across all synthetic benchmarks, SigMA outperformed CNN, LSTM, vanilla Transformer, and Deep Signature baselines in terms of accuracy, robustness, and parameter efficiency.

The empirical case studies further highlight the practical relevance of the proposed approach, particularly its ability to overcome the distribution mismatch between simulated training data and real-world observations. 
In the calibration of equity-index realized volatility, SigMA provided accurate and stable estimates of the Hurst (Hölder) exponent with significantly lower model complexity than competing architectures. In the analysis of Li-ion battery degradation data, SigMA captured persistent long-memory effects more reliably than conventional deep learning models, producing stable estimates that align well with classical statistical benchmarks while being less sensitive to non-stationarity and short-window artifacts. 
This generalizability likely stems from signature representations capturing the structural properties of trajectories rather than specific marginal distributions. 

From a broader perspective, this work illustrates how signature-based representations and attention mechanisms can be effectively combined for parameter inference in non-Markovian stochastic systems. While our focus has been on fBm-driven models, the methodology is not tied to a specific stochastic process and may be extended to other classes of path-dependent dynamics. 

While SigMA generally works well, it struggles at theoretical boundaries of the Hurst parameter. 
As $H\to0$, paths become extremely rough, and the key details shift to higher-order signature terms. 
By defaulting to $N=3$, SigMA loses this information and reduces in accuracy, while increasing $N$ is too computationally expensive. 
At the other extreme ($H\approx0.5$), the process is memoryless. 
SigMA's self-attention mechanism isn't needed here and may actually overfit by searching for nonexistent long-term dependencies. 
For such cases, simpler classical estimators are a better choice. 

A number of potential directions for future research arise from this work. In particular, extending the proposed framework to multivariate and coupled stochastic systems would be of interest, especially in high-dimensional finance and engineering applications. 
In addition, incorporating uncertainty quantification such as predictive intervals or Bayesian variants of SigMA could enhance interpretability.

\paragraph{Code and Data Availability}
The code used in this work is publicly available \href{https://github.com/changanluoxue/SigMA.git}{[GitHub link]}.

\appendix
\section{Architectural details of baseline and signature-based neural networks}
\label{apdix-architecture}
This appendix provides detailed descriptions of all neural network architectures used throughout the numerical experiments, including both baseline benchmark models (CNN, LSTM, Transformer, DeepSigNet) and the proposed signature-based variants (SigMA's ablations). 
\begin{itemize}
	\item The architecture of CNN model:
	\begin{itemize}[noitemsep]
		\item[-] a convolutional layer with 32 channels and kernel size 20;
		\item[-] the leaky ReLu transform with negative slope 0.1;
		\item[-] a max pooling layer with size 3;
		\item[-] a dropout layer with rate 0.25;			
		\item[-] a convolutional layer with 64 channels and kernel size 20;
		\item[-] the leaky ReLu transform with negative slope 0.3;
		\item[-] a max pooling layer with size 3;
		\item[-] a dropout layer with rate 0.25;
		\item[-] a convolutional layer with 128 channels and kernel size 20;
		\item[-] the leaky ReLu transform with negative slope 0.1;
		\item[-] a max pooling layer with size 3;
		\item[-] a dropout layer with rate 0.4;
		\item[-] a dense layer with 128 units;
		\item[-] the leaky ReLu transform with negative slope 0.1;
		\item[-] a dropout layer with rate 0.3;
		\item[-] a dense layer with 1 units;
		\item[-] a non-linear transform with sigmoid function.			
	\end{itemize}
	\item The architecture of DeepSigNet model:
	\begin{itemize}[noitemsep]
		\item[-] a convolutional layer with 3 channels and kernel size 3;
		\item[-] augmentation with time and original values;
		\item[-] a signature transform with truncation order $N=3$;
		\item[-] a fully connected feedforward neural network with 5 hidden layers, each of size 32 and ReLU activation;
		\item[-] a non-linear transform with sigmoid function.
	\end{itemize}
	\item The architecture of Transformer model:
	\begin{itemize}[noitemsep]
		\item[-] a convolutional layer with 153 channels and kernel size 3;
		\item[-] augmentation with time and original values;
		\item[-] a multi-head self-attention layer with the number of heads $h=N=3$;
		\item[-] a fully connected feedforward neural network with 5 hidden layers, each of size 32 and ReLU activation;
		\item[-] a non-linear transform with sigmoid function.
	\end{itemize}
	\item The architecture of LSTM model:
	\begin{itemize}[noitemsep]
		\item [-] a standardizing layer to transform the sequence to its increments;
		\item [-] a standardization on the increments;
		\item [-] a two-layer stacked LSTM with 128 hidden units per layer;
		\item [-] a fully connected feedforward neural network with 2 hidden layers of sizes 128 and 64, and PReLU activation.
	\end{itemize}
\end{itemize}

\begin{itemize}
	\item The architecture of SigSA model:
	\begin{itemize}[noitemsep]
		\item[-] augmentation with time;
		\item[-] a signature transform with truncation order $N=3$;
		\item[-] a single-head self-attention layer;
		\item[-] a non-linear transform with sigmoid function.
	\end{itemize}
	\item The architecture of SigMA without convolutional layer:
	\begin{itemize}[noitemsep]
		\item[-] augmented with time;
		\item[-] taking the signature transform with truncation order $N=3$;
		\item[-] the multi-head self-attention layer with the number of heads $h=N=3$ and each different head corresponds to a different signature level;
		\item[-] a fully connected feedforward neural network with 5 hidden layers, each of size 32 and ReLU activation;			
		\item[-] a non-linear transform with sigmoid function.			
	\end{itemize}
	\item The architecture of SigMA without MLP:
	\begin{itemize}[noitemsep]
		\item[-] a convolutional layer with 3 channels and kernel size 3;
		\item[-] augmentation with time and original values;
		\item[-] a signature transform with truncation order $N=3$;
		\item[-] a multi-head self-attention layer with the number of heads $h=N=3$ and each different head corresponds to a different signature level;
		\item[-] a linear transform to appropriate dimension.
	\end{itemize}
	\item The architecture of SigMA without convolutional layer \& MLP:
	\begin{itemize}[noitemsep]
		\item[-] augmentation with time;
		\item[-] a signature transform with truncation order $N=3$;
		\item[-] a multi-head self-attention layer with the number of heads $h=N=3$ and each different head corresponds to a different signature level;					\item[-] a linear transform to appropriate dimension.
	\end{itemize}
\end{itemize}

\section{Hyperparameter grid search for convolutional and MLP layers}
\label{apdix-sensitivity-structure-hyperparameter}
\begin{table}[H]
	\centering
	\rotatebox{90}{
		\begin{minipage}{0.9\textheight}
			\caption{Test RMSEs for Hurst estimation across stochastic processes with length 100, obtained using SigMA under different combinations of convolutional kernel sizes and MLP hidden layer sizes. Results are averaged over 3 training runs. Here, $H\sim\text{Uniform}(0,\,1)$ for the fBm, $H\sim\text{Uniform}(0.5,\,1)$ for the fOU, and $H\sim\text{Beta}(1,\,9)$ for the rHeston. The default configuration is marked. }
			\label{tab-sensitivity-structure-hyperparameter}
			\smallskip
			\centering
			\renewcommand\arraystretch{1.5}
			\setlength{\tabcolsep}{6pt}
			\begin{tabular}{l c c c c c}
				\toprule
				\multirow{2}{*}{Convolutional kernel size} & \multirow{2}{*}{MLP hidden layer size} & \multicolumn{3}{c}{Stochastic processes} & \multirow{2}{*}{$\#$Params} \\
				\cline{3-5}
				& &fBm&fOU&rHeston& \\
				\hline
				\multirow{3}{*}{2}
				&16&1.68e-02&3.29e-02&5.58e-02&79095\\
				&32&1.69e-02&3.14e-02&5.40e-02&87223\\
				&64&1.76e-02&2.93e-02&5.32e-02&109623\\
				\hline
				\multirow{3}{*}{3 (Default)}
				&16&1.57e-02&2.82e-02&5.51e-02&79098\\
				&32 (Default)&1.49e-02&2.70e-02&5.01e-02&87226\\
				&64&1.38e-02&2.55e-02&4.56e-02&109626\\
				\hline
				\multirow{3}{*}{4}
				&16&1.55e-02&2.64e-02&5.64e-02&79101\\
				&32&1.47e-02&2.72e-02&5.30e-02&87229\\
				&64&1.60e-02&3.09e-02&3.99e-02&109629\\
				\bottomrule	
			\end{tabular}
		\end{minipage}}
\end{table}

\section{Individual parameter results for joint multi-parameter estimation}
\label{apdix-multiple-individual}
\begin{table}[H]
	\centering
	\rotatebox{90}{
		\begin{minipage}{0.9\textheight}
			\caption{Test results for multiple parameters estimation across stochastic processes with length 500, obtained using various NN models, are based on $10$ training runs. Here, numbers in parentheses represent the variance of the RSEs.  The best accuracies across different NN models are in bold. }
			\label{tab-multiple2}
			\smallskip
			\centering
			\renewcommand\arraystretch{1.5}
			\setlength{\tabcolsep}{6pt}
			\begin{tabular}{l c c c c c c c c c}
				\toprule
				\multirow{2}{*}{Models}&\multicolumn{4}{c}{RMSEs for fOU}&\multicolumn{4}{c}{RMSEs for rHeston}&\multirow{2}{*}{$\#$Params}\\
				\cline{2-9}
				&$H$&$\alpha$&$\mu$&$\sigma$&$H$&$\kappa_1$&$\kappa_2$&$\theta$&\\
				\hline
				\multirow{2}{*}{Transformer}&0.165&0.764&0.149&0.664&0.086&0.993&0.130&0.416&\multirow{2}{*}{2558005}\\
				&{\scriptsize(1.12e-2)}&{\scriptsize(2.96e-1)}&{\scriptsize(8.28e-3)}&{\scriptsize(1.49e-1)}&{\scriptsize(3.27e-3)}&{\scriptsize(3.71e-1)}&{\scriptsize(4.86e-3)}&{\scriptsize(8.71e-2)}&\\
				\multirow{2}{*}{SigMA}&\textbf{0.120}&\textbf{0.761}&0.139&\textbf{0.643}&\textbf{0.071}&\textbf{0.970}&0.110&\textbf{0.412}&\multirow{2}{*}{87341}\\
				&{\scriptsize(5.38e-3)}&{\scriptsize(2.51e-1)}&{\scriptsize(5.75e-3)}&{\scriptsize(1.56e-1)}&{\scriptsize(3.75e-3)}&{\scriptsize(3.38e-1)}&{\scriptsize(4.03e-3)}&{\scriptsize(8.40e-2)}&\\
				\multirow{2}{*}{DeepSigNet}&0.451&1.228&0.741&0.998&0.084&0.992&0.100&0.414&\multirow{2}{*}{9360}\\
				&{\scriptsize(2.25e-1)}&{\scriptsize(9.92e-1)}&{\scriptsize(7.50e-1)}&{\scriptsize(4.63e-1)}&{\scriptsize(3.16e-3)}&{\scriptsize(3.57e-1)}&{\scriptsize(3.67e-3)}&{\scriptsize(8.09e-2)}&\\
				\multirow{2}{*}{CNN}&0.129&1.051&\textbf{0.128}&0.648&0.078&1.241&\textbf{0.088}&0.587&\multirow{2}{*}{501220}\\
				&{\scriptsize(5.13e-3)}&{\scriptsize(3.28e-1)}&{\scriptsize(4.60e-3)}&{\scriptsize(1.64e-1)}&{\scriptsize(2.69e-3)}&{\scriptsize(5.62e-1)}&{\scriptsize(2.56e-3)}&{\scriptsize(1.29e-1)}&\\
				\multirow{2}{*}{LSTM}&0.158&0.977&0.212&1.030&0.107&1.275&0.157&0.909&\multirow{2}{*}{8383299}\\
				&{\scriptsize(9.33e-3)}&{\scriptsize(4.43e-1)}&{\scriptsize(1.70e-2)}&{\scriptsize(3.70e-1)}&{\scriptsize(4.67e-3)}&{\scriptsize(5.74e-1)}&{\scriptsize(9.51e-3)}&{\scriptsize(2.82e-1)}&\\
				\bottomrule
			\end{tabular}
		\end{minipage}}
\end{table}

\bibliographystyle{elsarticle-num-names}
\bibliography{bib}

\end{document}